\documentclass[10pt,twocolumn,letterpaper]{article}

\usepackage{cvpr}              

\usepackage{subcaption}
\usepackage[ruled,vlined]{algorithm2e}
\usepackage{colortbl}
\usepackage{wrapfig}

\definecolor{cvprblue}{rgb}{0.21,0.49,0.74}
\usepackage[pagebackref,breaklinks,colorlinks,allcolors=cvprblue]{hyperref}

\def\paperID{16792} 
\def\confName{CVPR}
\def\confYear{2025}

\title{Perturb-and-Revise: Flexible 3D Editing with Generative Trajectories}

\newcommand*{\affaddr}[1]{#1}
\newcommand*{\affmark}[1][*]{\textsuperscript{#1}}

\author{
Susung Hong\affmark[1]\quad Johanna Karras\affmark[1]\quad Ricardo Martin-Brualla\affmark[2]\quad Ira Kemelmacher-Shlizerman\affmark[1,2]\\
\\
\affaddr{\affmark[1]University of Washington}\quad \affaddr{\affmark[2]Google Research}\\
\vspace{-1cm}
}

\begin{document}

\twocolumn[{
\renewcommand\twocolumn[1][]{#1}
\maketitle
\begin{center}
\centering
\captionsetup{type=figure}
\includegraphics[width=0.90\textwidth]{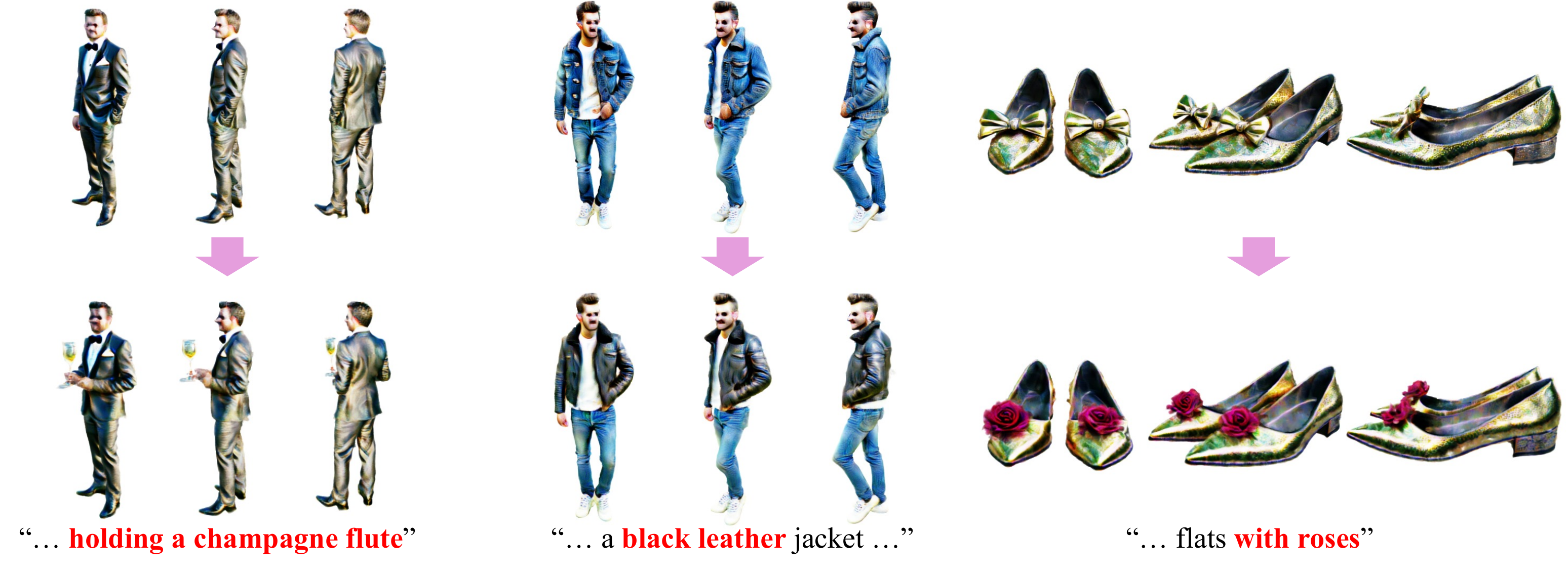}
\vspace{-12pt}
\caption{\textbf{Perturb-and-Revise} takes a source NeRF and an edit prompt as input and produces the edited result through: (1) versatile initialization via parameter perturbation, (2) multi-view consistent score distillation, and (3) refinement with the identity-preserving gradient.}
\label{fig:teaser}
\end{center}%
}]

\maketitle
\begin{abstract}

Recent advancements in text-based diffusion models have accelerated progress in 3D reconstruction and text-based 3D editing. Although existing 3D editing methods excel at modifying color, texture, and style, they struggle with extensive geometric or appearance changes, thus limiting their applications. To this end, we propose \textbf{Perturb-and-Revise}, which makes possible a variety of NeRF editing. First, we \textbf{perturb} the NeRF parameters with random initializations to create a versatile initialization. The level of perturbation is determined automatically through analysis of the local loss landscape. Then, we \textbf{revise} the edited NeRF via generative trajectories. Combined with the generative process, we impose identity-preserving gradients to refine the edited NeRF. Extensive experiments demonstrate that Perturb-and-Revise facilitates flexible, effective, and consistent editing of color, appearance, and geometry in 3D. For 360° results, please visit our project page: \href{https://susunghong.github.io/Perturb-and-Revise}{https://susunghong.github.io/Perturb-and-Revise}.

\end{abstract}

\begin{figure*}[t]
\centering
\includegraphics[width=1.0\linewidth]{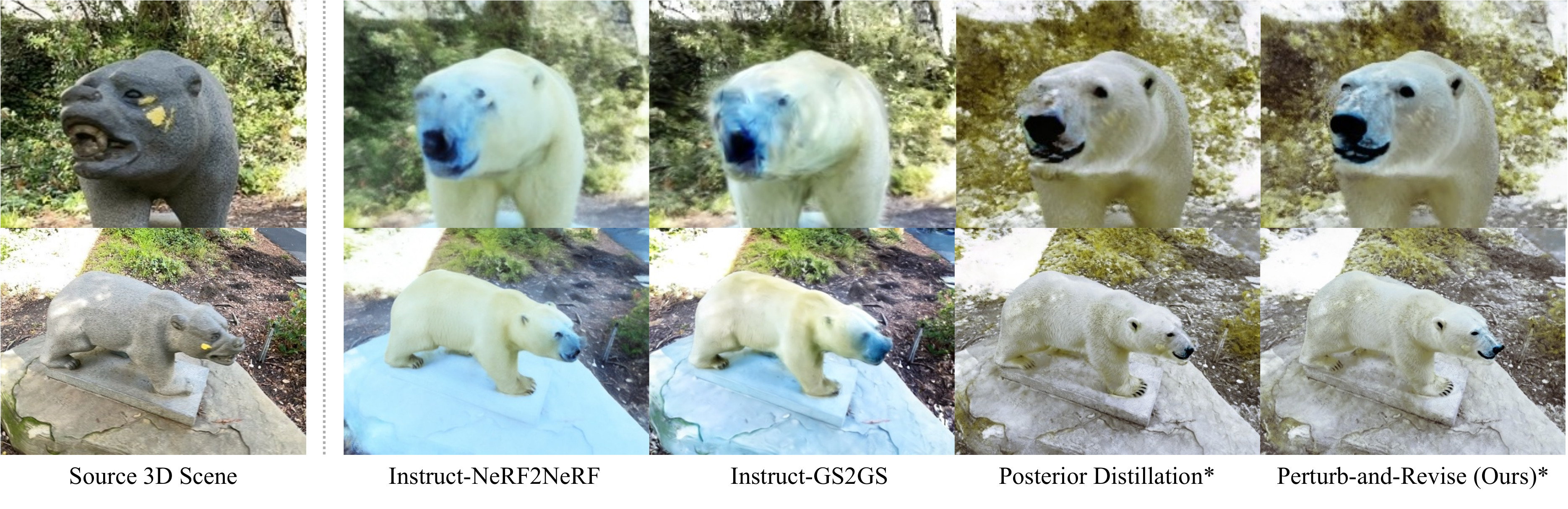}
\vspace{-27pt}
\caption{We qualitatively compare our method to Instruct-NeRF2NeRF~\cite{haque2023instruct}, Instruct-GS2GS~\cite{igs2gs}, and Posterior Distillation~\cite{koo2024posterior}. In this task, the goal is to edit the source ``bear" into a ``polar bear." Compared to related works, our approach reconstructs a more realistic face and better overall geometry. An asterisk (*) denotes that we use an identical update rule and schedule.}
\label{fig:teaser-bear}
\vspace{-15pt}
\end{figure*}

\section{Introduction}

Neural Radiance Fields (NeRFs)~\cite{mildenhall2021nerf} have revolutionized the creation of high-quality 3D scenes, marking a significant advancement in 3D reconstruction technology. Beyond their initial applications, NeRFs have enabled flexible generation of 3D content from models trained solely on image-text pairs~\cite{jain2022zero,poole2022dreamfusion} through score distillation~\cite{poole2022dreamfusion,wang2023score,koo2024posterior}.

However, editing 3D content---an essential aspect of refinement and customization---remains a time-consuming and labor-intensive process across various industries, including animation, manufacturing, design, and gaming. This challenge is particularly pronounced with NeRFs, where color and density attributes are intricately encoded within their parameters. Consequently, there is an ongoing need for more intuitive and universally accessible tools for editing NeRFs, leveraging user-friendly interfaces such as text prompts for broader applicability and faster production.

Fortunately, recent innovations in text-conditioned diffusion models have enabled significant advances in 3D editing using natural language prompts. For instance, leveraging diffusion models~\cite{brooks2022instructpix2pix,rombach2022high}, recent works such as Instruct-NeRF2NeRF~\cite{haque2023instruct} and Posterior Distillation~\cite{koo2024posterior} propose methods that use text prompts to modify 3D scenes. While these methods excel in altering the color or style of an object, they struggle with edits involving significant geometric or appearance changes, limiting their applications.

\begin{figure}[t]
\centering
\includegraphics[width=0.9\linewidth]{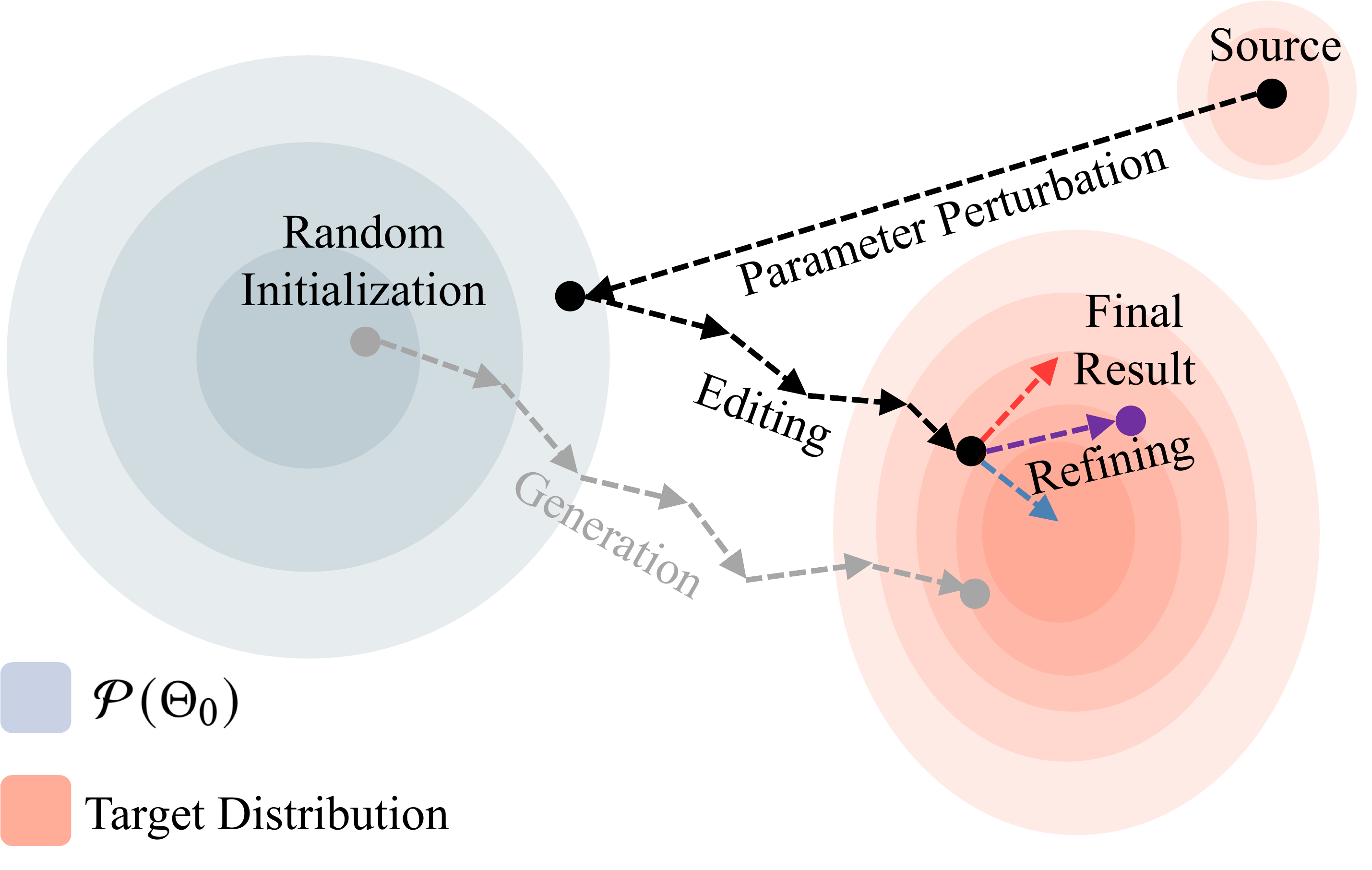}
\vspace{-10pt}
\caption{Conceptual figure. The target distribution in the figures represents the conditional distribution of NeRF parameters relative to the edit prompt, and  $\mathcal{P}(\Theta)$ denotes the distribution of randomly initialized NeRF parameters. First, parameter perturbation enables the parameters to escape from local minima and follow a natural generative path. Subsequently, during the refining process, the tug-of-war between two vectors, $\lambda_{\text{d}}\nabla_{\theta}d(\theta_\tau, \theta_\textrm{src})$ (the red arrow) and $d\theta_\tau$ (the blue arrow), pushes the actual parameters into a region that is closer to either the source parameters or the high-density region specified by the edit prompt, following the purple arrow.}
\label{fig:conceptual}
\vspace{-15pt}
\end{figure}

To overcome these limitations, we propose a 3D object editing framework called \textbf{Perturb-and-Revise (PnR)}, which leverages existing 3D generation methods for editing in a flexible and natural way (Figs.~\ref{fig:teaser} and \ref{fig:teaser-bear}). Drawing on the notion that NeRF parameters optimized with score distillation can be considered as a particle~\cite{wang2024prolificdreamer} or data point~\cite{dupont2022data}, we propose a novel method that leverages perturbation at the parameter level to perform text-based edits requiring various changes. Specifically, we construct versatile NeRF initialization by interpolating the source NeRF and a random NeRF initialization. Intuitively, perturbation in the parameter space helps the particle escape local minima and facilitates its following of the natural trajectory of the generative ODE towards the distribution of the desired edit (Fig.~\ref{fig:conceptual}), allowing for a wide range of challenging edits including changing the pose and introducing new objects.

Additionally, to determine the amount of perturbation needed to escape the basin of attraction without costly searching algorithms, we propose an algorithm based on the loss landscape near the source parameters, by simulating a few score distillation steps ahead of the parameter perturbation and optimization. After the early step of our framework, to achieve more similarity to the original object and refine the quality of the result, we employ the Identity-Preserving Gradient (IPG) to increase the fidelity of the edited NeRF to the source NeRF. As a result, the output closely resembles the source while still maintaining fidelity to the intended edits.

We extensively evaluate our method on 3D fashion objects, as well as general objects in Objaverse~\cite{deitke2023objaverse}, on a variety of appearance- and geometry-based edits. Furthermore, we demonstrate that our method achieves state-of-the-art results on various 3D editing baselines.

\section{Related Work}

\paragraph{Diffusion models.} Diffusion models~\cite{ho2020denoising,song2021denoising}, which are closely associated with score-based models~\cite{song2020score,song2019generative,karras2022elucidating}, have recently shown remarkable sample quality in image synthesis. Latent Diffusion Models (LDM)~\cite{rombach2022high}, which perform the diffusion forward and backward process in the latent space~\cite{esser2021taming}, have demonstrated their efficacy and generation quality. This trend has been scaled up by training diffusion models on large-scale text-image datasets~\cite{schuhmann2022laion} for text-conditional generation. The prosperity of text-to-image diffusion models has led to their application in complex content creation, such as 3D scenes and videos~\cite{khachatryan2023text2video,poole2022dreamfusion,wang2024prolificdreamer,hong2023debiasing,hong2023direct2v,huang2023free,wang2023score,metzer2022latent,lin2023magic3d,wu2022tune,ho2022imagen,seo2024retrieval,blattmann2023align}. Recently, many adapting methods of text-to-image diffusion models have been proposed to achieve either generation with additional conditions~\cite{zhang2023adding,mou2023t2i,karras2023dreampose,hong2023improving,brooks2022instructpix2pix,zhu2023tryondiffusion,hong2024smoothed} or to maintain consistency needed for videos~\cite{qi2023fatezero,wu2022tune,khachatryan2023text2video,hong2023direct2v} or 3D scenes~\cite{shi2023mvdream,riu2023zero}.

\vspace{-10pt}
\paragraph{Score distillation.} Score distillation~\cite{poole2022dreamfusion,wang2023score,wang2024prolificdreamer}, a recent trend in synthesizing underlying parameterizations, mostly in 3D scene generation, uses only score-based models~\cite{rombach2022high,saharia2022photorealistic} operating on 2D data. This approach has revolutionized text-to-3D generation by introducing an elegant method that does not require intense training on a 3D dataset. It enables the rich and flexible knowledge of 2D text-to-image models to be transferred to 3D  generation~\cite{metzer2022latent,lin2023magic3d,wang2023score,hong2023debiasing,wang2024prolificdreamer}. However, due to bias in 2D models, a multi-view consistency problem has arisen, necessitating further analysis and generalization of the original method~\cite{hong2023debiasing,shi2023mvdream,riu2023zero}. Another limitation of score distillation is its time-consuming optimization, with the controllability and editing capability remaining underexplored.

\vspace{-10pt}
\paragraph{Text-based NeRF editing.}
In recent years, neural radiance fields (NeRF)~\cite{mildenhall2021nerf} have emerged as a groundbreaking approach for generating photorealistic novel views of a scene captured in photographs, and have been extended in many follow-up works~\cite{kerbl20233d,muller2022instant,martin2021nerf,park2021nerfies}. In response to these advancements, several techniques for modifying NeRF have been proposed. Traditional methodologies for editing NeRF include the alteration of materials and lighting~\cite{srinivasan2021nerv,verbin2022ref,munkberg2022extracting}, as well as the spatial manipulation of objects via bounding box frameworks~\cite{ost2021neural,yu2021unsupervised}. Additionally, there has been exploration in the stylization of NeRFs, including EditNeRF~\cite{liu2021editing}, Clip-NeRF~\cite{wang2022clip}, and NeRF-Art~\cite{wang2023nerf}, as well as in distilling 2D features into radiance fields, such as in Distilled Feature Fields~\cite{kobayashi2022decomposing} and Neural Feature Fusion Fields~\cite{tschernezki2022neural}, which enables nuanced, guided edits based on language or images. Crucially, Instruct-NeRF2NeRF~\cite{haque2023instruct} focuses on user-friendly, language-based editing commands, using an instruction-driven, 2D image-conditioned diffusion model~\cite{brooks2022instructpix2pix} for more intuitive and context-aware 3D editing. However, even in a trend using 2D language-based models such as CLIP~\cite{radford2021learning} or text-to-image diffusion models~\cite{brooks2022instructpix2pix,rombach2022high,saharia2022photorealistic}, these methodologies are unable to control large geometrical changes, such as changing the pose or introducing a massive object. Posterior Distillation~\cite{koo2024posterior}, instead of naively matching the noise, proposes a way to use a loss that matches the stochastic latents~\cite{huberman2024edit,wu2023latent} of source and target images. Still, this method falls into issues such as slower convergence and large computational costs.

\section{Background}
\label{sec:background}

Diffusion models are a type of generative model that iteratively restores an image from Gaussian noise. The training objective of diffusion models involves training a neural network to estimate the score for a noised image $y$ given the noise level $\sigma$ via denoising score matching~\cite{hyvarinen2005estimation}. Following the preconditioning convention~\cite{karras2022elucidating}, a diffusion model, parameterized by $\phi$, can be interpreted as a denoiser $D_\phi(y; \sigma)$, which minimizes the weighted L2 loss across different $\sigma$ values in data samples.

In the context of text-based 3D scene generation, Score Distillation Sampling (SDS)~\cite{poole2022dreamfusion} updates the 3D scene representation given a text caption $c$ using the following rule:
\begin{align}
\begin{split}
&\nabla_\theta \mathcal{L}_{\textrm{SDS}}(\phi, c, x = g(\theta, \psi)) = \\&-\mathbb{E}_{\sigma \sim \Sigma, n \sim \mathcal{N}(0,\sigma^2 I)}\Big[\omega(\sigma)\big(D_\phi(x + n; \sigma, c) - x \big)\frac{\partial x}{\partial \theta}\Big].
\end{split}
\end{align}
In this equation, $g(\cdot)$ is the differentiable renderer, $\psi$ is the random camera pose, $\Sigma$ is a predefined distribution from which the noise level for the denoiser is sampled, and $\omega(\cdot)$ denotes the weighting function.

SDS can be viewed as a form of particle-based variational inference~\cite{chen2018unified,liua2022geometry,wang2019function,dong2022particle,liu2016stein,wang2024prolificdreamer}. In this context, the update rule can be expressed as a generative ODE over an optimization step $\tau$, derived via Wasserstein gradient flow~\cite{chen2018unified}:
\begin{equation}
\frac{d\theta_\tau}{d\tau}=-\mathbb{E}_{\sigma \sim \Sigma, n \sim \mathcal{N}(0,\sigma^2 I)}\Big[\omega(\sigma)\big(D_\phi(x + n; \sigma, c) - x \big)\frac{\partial x}{\partial \theta_\tau}\Big].
\label{eq:ode}
\end{equation}
For further details, please see the supplementary material.

As background, multi-view diffusion models~\cite{shi2023mvdream}, denoted as $\{D^{\textrm{M}}_\phi(g(\theta, \psi_i) + n; \sigma, c, \psi_i)\}_{i=1}^{N}$, consistently generate multi-view images simultaneously. Here, $N$ denotes the total number of multi-views generated simultaneously, and $\psi_i$ denotes the $i$-th viewpoint. Note that for $N=1$, these are ordinary diffusion models. For $N>1$, they are implemented by adapting the attention module to attend to different frames~\cite{shi2023mvdream} and by training on 3D assets~\cite{deitke2023objaverse}.

\section{Method}
In text-based editing scenarios, we assume that a user provides a text prompt, $c_\textrm{edit}$, and a source 3D object represented by the parameterization $\theta_\textrm{src}$. The aim is to modify the 3D object into a new form $\theta^* = \mathcal{E}(\theta_\textrm{src}, c_\textrm{edit})$, ensuring that the transformation aligns with the text prompt and retains a resemblance to the original 3D object.

At a high level, we leverage the generative ODE of 3D objects through novel parameter perturbation while preserving proximity to the source object. In Sec.~\ref{sec:parameter-perturbation}, we introduce perturbation in parameter space that enables the parameters to escape local minima and converge to natural generative trajectories. In Sec.~\ref{sec:eta}, we propose a novel algorithm to determine the degree of perturbation by analyzing the local loss landscape. Finally, in Sec.~\ref{sec:correction}, we propose balancing fidelity between the text prompt and the source object to place the final output in the desired region.

\subsection{Parameter Perturbation for Flexible Object Editing}
\label{sec:parameter-perturbation}

\begin{figure}[t]
\centering
\includegraphics[width=1.0\linewidth]{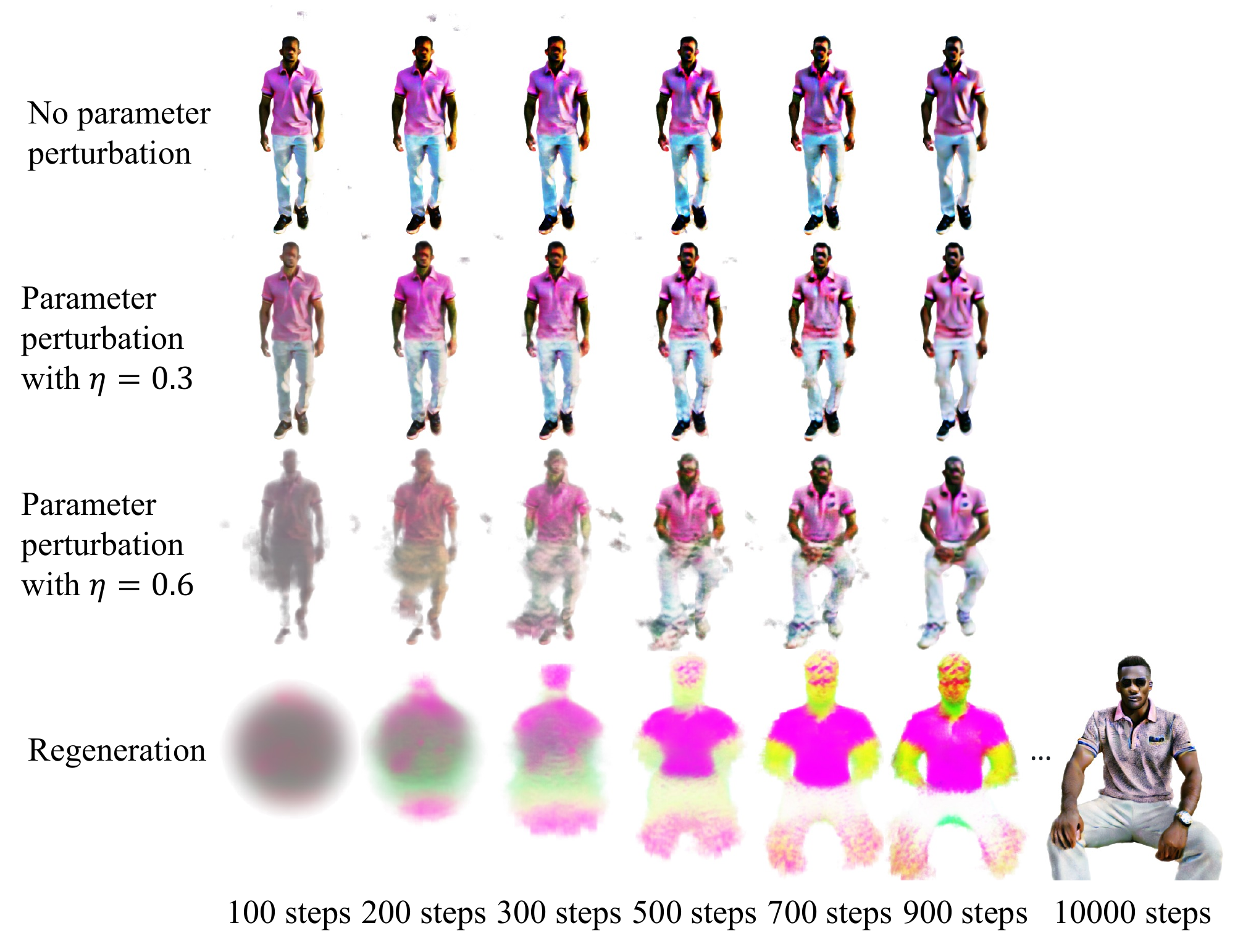}
\vspace{-25pt}
\caption{Effect of parameter perturbation. In this example, we aim to make a NeRF model of a standing person sit down using the word ``sitting." The scene converges quickly even with large perturbations ($\eta = 0.6$), while complete regeneration yields blurry rendering results given the same number of optimization steps.}
\label{fig:fashion-timestep}
\vspace{-15pt}
\end{figure}

Given the generative ODE in Eq.~\ref{eq:ode}, which minimizes the energy functional between distributions, a fully-optimized NeRF of a 3D object is considered to exhibit relatively low energy. This state corresponds to the particle residing in a local minimum, which remains stable despite alterations in the text prompts, as it is still capable of producing realistic renderings. As exemplified in the first row of Fig.~\ref{fig:fashion-timestep}, naively using $\theta_{\text{src}}$ as initialization and making independent predictions on 2D noisy images with diffusion models are insufficient to handle the consistent changes in overall appearance and geometry that occur in a 3D object.

Interpreting neural fields as a particle~\cite{wang2024prolificdreamer} or a data point~\cite{dupont2022data}, we propose parameter-level perturbation for an editing task, which facilitates changes in $\theta_{\text{src}}$ by enabling the editing particle ODE to hijack the natural, coarse-to-fine generative process in Eq.~\ref{eq:ode} that the parameters are likely to follow. In other words, since adding more noise to the parameters is equivalent to undoing more of the optimization process, this approach enables the parameters to be flexible, allowing for significant changes.

Specifically, let $\mathcal{P}(\Theta_0)$ denote the probability density function over $\Theta_0$, which is a set of random parameterizations. Further, assume that we have a fully optimized 3D scene, $\theta_{\mathrm{src}}$. Then, we obtain the perturbed version of $\theta_{\mathrm{src}}$ by using linear interpolation:
\begin{equation}
\theta_\textrm{perturbed} = \textrm{Lerp}(\theta_\textrm{src}, \theta_0, \eta) \quad \textrm{where} \quad \theta_0 \sim \mathcal{P}(\Theta_0).
\label{eq:parameter-perturbation}
\end{equation}
where $\eta \in [0,1]$ denotes the perturbation amount and $\textrm{Lerp}(\theta_\textrm{src}, \theta_0, \eta) = (1-\eta) \cdot \theta_\textrm{src} + \eta \cdot \theta_\textrm{0}$. Note that $\eta=1$ indicates $\theta_\textrm{perturbed}$ is the random parameters. Theoretically, the interpolation makes the distribution of perturbed particles similar to the initializing distribution of NeRF, making it versatile. See the supplementary material for details.

Subsequently, we run the parameter ODE for textual editing, starting from $\theta_\textrm{perturbed}$, using a new prompt $c_\textrm{edit}$. However, we observe that adopting ordinary single-view diffusion models (when $N=1$) poses ambiguity when an edit introduces asymmetry or adds a new object. Therefore, we execute multi-view consistent updates as follows:
\begin{align}
\begin{split}
\frac{d\theta_\tau}{d\tau} = -\frac{1}{N}\sum^{N}_{i=1}\mathbb{E}_{\sigma, n}\Big[\big(D^{\textrm{M}}_\phi(z_i + n; \sigma, c_\textrm{edit}, \psi_i) - z_i \big)\frac{\partial z_i}{\partial \theta_\tau}\Big],
\label{eq:mv-ode}
\end{split}
\end{align}
where we define $z_i := g(\theta_\tau, \psi_i)$ and omit the weighting factor for brevity.

The conceptual illustration of parameter perturbation can be found in Fig.~\ref{fig:conceptual}. This approach essentially reverts the optimization process by interpolating between the initial state and the fully optimized state. Consequently, the particles are relocated to a less-optimized state characterized by a more moderate energy landscape. This repositioning allows the particle ODE (Eq.~\ref{eq:mv-ode}) to intervene in the generative process with a new text prompt $c_\textrm{edit}$, guiding the parameter trajectory. As shown in Fig.~\ref{fig:fashion-timestep}, increasing the noise level in the parameters inversely correlates with the degree of reversion of the optimization, allowing for more radical modifications in the model outputs.

\subsection{Determining \texorpdfstring{$\eta$}{} by Analyzing Loss Landscape}
\label{sec:eta}

In Sec.~\ref{sec:parameter-perturbation}, we see that $\eta$ controls parameter versatility by determining how similar the distribution of parameters is to the distribution of random NeRF initializations. The required degree of versatility depends on both the type of edits and the current source NeRF parameters. However, finding an optimal $\eta$ through standard mechanisms, such as grid or random search, is computationally expensive. Therefore, we propose a novel algorithm that explores the basin of attraction surrounding local minima to determine $\eta$ by analyzing the loss landscape.

Specifically, we leverage the loss function as a proxy to measure the depth and volume of the basin of attraction. Interestingly, for some examples, we observe that the loss function even increases during optimization steps, supporting our claim in Sec.~\ref{sec:parameter-perturbation} that the parameters are in a low-energy state. To address this, we simulate several optimization steps with $c_\textrm{edit}$ and calculate the total loss decrease prior to parameter perturbation. For algorithmic stability, we compute the difference between the averages of the first few steps and the last few steps. We then determine $\eta(\theta_\textrm{src}, c_\textrm{edit})$ based on the source parameters $\theta_\textrm{src}$ and the text prompt $c_\textrm{edit}$ using an inverted exponential decay function. Subsequently, we apply parameter perturbation in Eq.~\ref{eq:parameter-perturbation} with $\eta = \eta(\theta_\textrm{src}, c_\textrm{edit})$. The complete algorithms for our parameter perturbation and $\eta$ selection are provided in the supplementary material.

\begin{figure*}[t]
\centering
\includegraphics[width=0.95\linewidth]{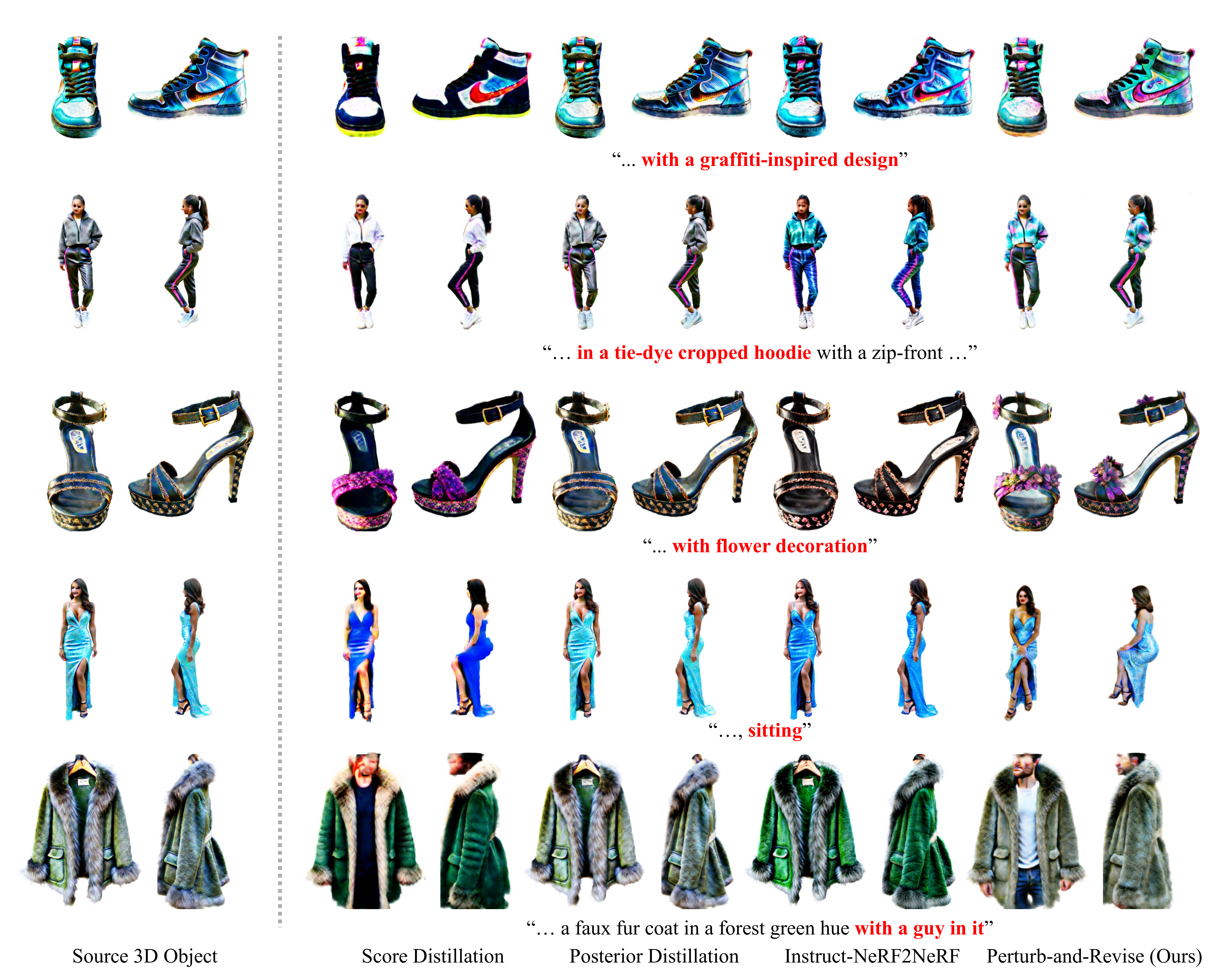}
\vspace{-15pt}
\caption{Baseline comparisons with a wide range of edits. We compare our method with Score Distillation Sampling (SDS)~\cite{poole2022dreamfusion}, Posterior Distillation Sampling (PDS)~\cite{koo2024posterior}, and Instruct-NeRF2NeRF~\cite{haque2023instruct}. For SDS and PDS, we use MVDream~\cite{shi2023mvdream} as the backbone for fair comparison. SDS alters the appearance and texture of the source objects and is unable to handle edits that require extensive geometric changes (3rd, 4th, and 5th rows). PDS is not capable of making significant edits and cannot deviate far from local minima due to its high preservation term from the start~\cite{koo2024posterior}. While Instruct-NeRF2NeRF changes the texture of objects as desired, it cannot address geometric changes. In contrast, our method is capable of various types of edits, including those involving large geometric changes.}
\label{fig:fashion-comparison}
\vspace{-15pt}
\end{figure*}

\begin{figure*}[t]
\centering
\includegraphics[width=0.95\linewidth]{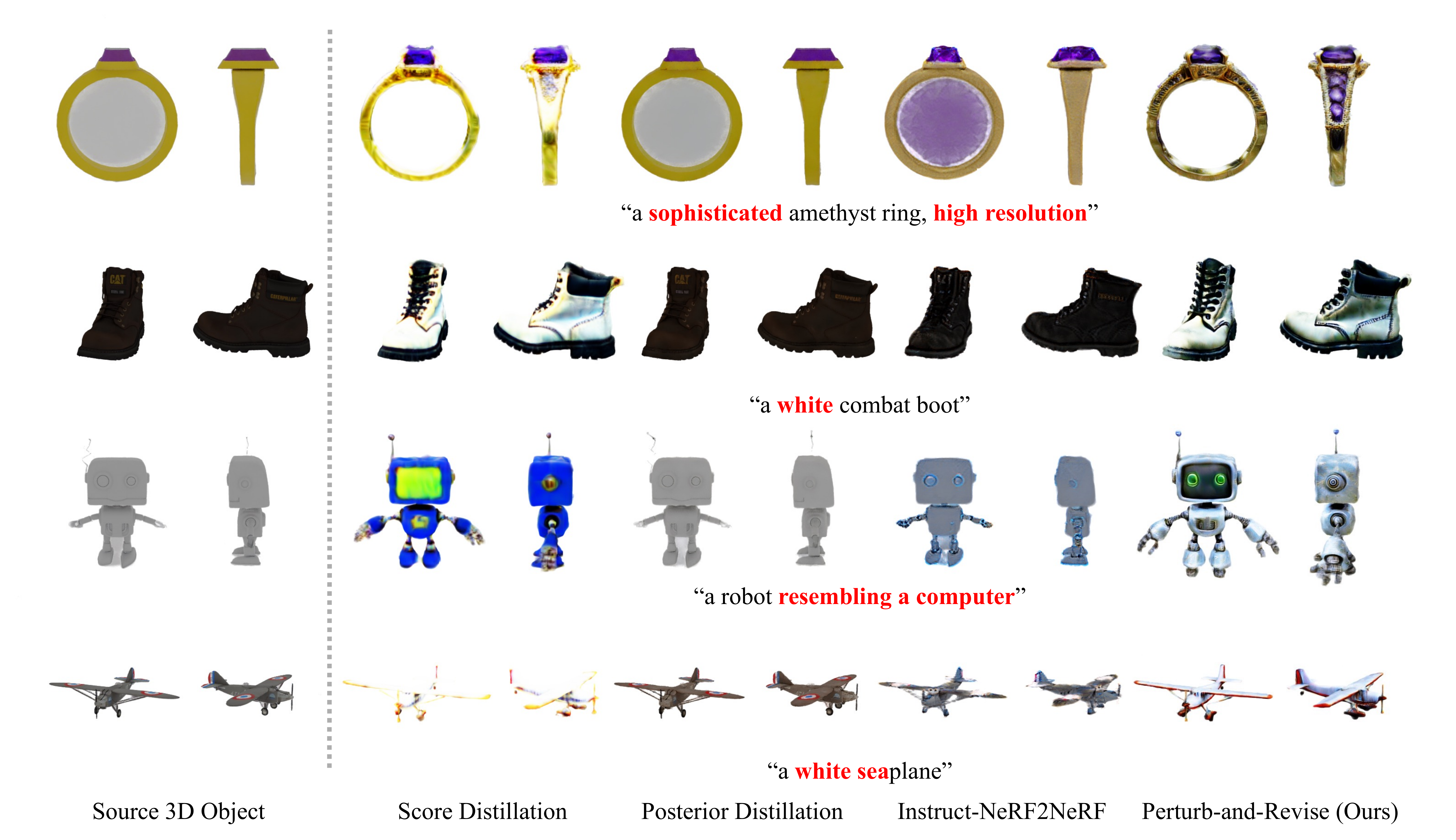}
\vspace{-15pt}
\caption{Baseline comparisons of editing various general 3D objects from the Objaverse dataset~\cite{deitke2023objaverse}.
}
\label{fig:objaverse-1}
\vspace{-5pt}
\end{figure*}

\subsection{Identity-Preserving Gradient (IPG)}
\label{sec:correction}

While our parameter perturbation in Sec.~\ref{sec:parameter-perturbation} successfully enables flexible changes, there are still some estimation errors or biases arising from the diffusion model during this phase. On the other hand, imposing constraints that make it similar to the source object (e.g., L2 distance) conflicts with the generative ODE, making the optimization challenging. To counteract the estimation error and to circumvent the conflict, we introduce the Identity-Preserving Gradient (IPG) term that is added at later steps to the editing gradient presented in Sec.~\ref{sec:parameter-perturbation}.

Initially, we assume that the ideal parameters are closer to the source parameters and the high-density region of the target distribution. Then, as conceptually shown in Fig.~\ref{fig:conceptual}, we instigate a tug-of-war between two vectors: one symbolizes the velocity consistently pointing towards the high-density region as dictated by the edit prompt, and the other symbolizes the velocity towards $\theta_{\textrm{src}}$. Specifically, the velocity in Eq.~\ref{eq:mv-ode} propels the particle towards the region with a high likelihood given the text $c_{\textrm{edit}}$, while the other gradient represents the pull towards the original NeRF parameterization $\theta_{\textrm{src}}$. This effectively corrects the shift caused by the error in the editing process, ensuring it resembles the original while maintaining the intended edits. To this end, we extend the optimization process and compute IPGs during these additional steps.

\begin{table*}[t]
\footnotesize
\centering
\begin{tabular}{lccccc}
\toprule
\textbf{Metric} & Score Distillation~\cite{poole2022dreamfusion} & Posterior Distillation~\cite{koo2024posterior} & Instruct-NeRF2NeRF~\cite{haque2023instruct} & Perturb-and-Revise (Ours) \\
\midrule
CLIP-Dir-Sim$_\textrm{ViT-B/32}$↑ & 0.0480 & \textcolor{gray}{0.0287} & \underline{0.0583} & \textbf{0.0594} \\
CLIP-Dir-Sim$_\textrm{ViT-B/16}$↑ & 0.0418 & \textcolor{gray}{0.0304} & \textbf{0.0549} & \underline{0.0534} \\
CLIP-Dir-Sim$_\textrm{ViT-L/14}$↑ & 0.0415 & \textcolor{gray}{0.0264} & \underline{0.0539} & \textbf{0.0567} \\
CLIP-Dir-Sim$_\textrm{averaged}$↑ & 0.0438 & \textcolor{gray}{0.0285} & \underline{0.0557} & \textbf{0.0565} \\
\midrule
LPIPS$_\textrm{vgg}$↓ & \textcolor{gray}{0.1273} & \textbf{0.0337} & 0.1065 & \underline{0.1060} \\
LPIPS$_\textrm{alex}$↓ & \textcolor{gray}{0.1533} & \textbf{0.0215} & 0.1112 & \underline{0.1034} \\
\bottomrule
\end{tabular}
\vspace{-7pt}
\caption{Comparison of different methods for fashion object editing. The best, second-best, and worst values are highlighted in \textbf{bold}, \underline{underlined}, and \textcolor{gray}{gray}, respectively. Values represent averages across all edit types and prompts. Our approach achieves a better trade-off between faithfulness to edit prompts and preservation of source 3D objects. While PDS exhibits lower LPIPS, it mostly generates edited objects that are nearly identical to the source, as evidenced by CLIP directional similarity.}
\label{tab:quantitative-comparison}
\vspace{-15pt}
\end{table*}

Formally, we combine the result of Eq.~\ref{eq:mv-ode} with the IPG to define a refinement step as follows:
\begin{align}
\begin{split}
d\theta_{\tau}^{\text{refine}} = d\theta_\tau + \lambda_{\text{d}}\nabla_{\theta}d(\theta_\tau, \theta_\textrm{src}),
\label{eq:refine-similarity}
\end{split}
\end{align}
where $d(\cdot, \cdot)$ is a similarity metric. In practice, we observe that the combination of $L_1$ and perceptual loss~\cite{johnson2016perceptual}, which is also the preferred choice in NeRF~\cite{mildenhall2021nerf} training, is more robust and less susceptible to noise than using $L_2$ loss. Thus, given a random camera pose $\psi$, we use a combination of L1 distance and perceptual similarity between the rendered images $g(\theta_\tau,\psi)$ and $g(\theta_\mathrm{src},\psi)$, weighted by $\lambda_{\text{L1}}$ and $\lambda_{\text{p}}$, respectively.

\subsection{Timestep Annealing}
\label{sec:timestep}
In text-based 3D generation, some recent works have revealed that using timestep annealing, which adjusts the noise level added to the 2D rendered images according to the global optimization step, boosts the quality of the results~\cite{wang2024prolificdreamer,shi2023mvdream}. In the editing task, while we use a smaller noise level at the start, we also identify the annealing method as crucial for maintaining the original quality of the 3D object while reducing blurry textures. To this end, we regard $\Sigma$ as a function that depends on the editing step $\tau$, $\Sigma(\tau)$, from which the noise level $\sigma$ is sampled. Specifically, for $\tau \leq \mathrm{T}$, where $\mathrm{T}$ is the final step of the schedule, $\Sigma(\tau)$ is gradually annealed from $\mathcal{U}(\sigma_\textrm{min}^0, \sigma_\textrm{max}^0)$ to $\mathcal{U}(\sigma_\textrm{min}^\mathrm{T}, \sigma_\textrm{max}^\mathrm{T})$. This approach initially facilitates low-frequency edits, followed by fine-grained ones.

\begin{figure*}[t]
\centering
\includegraphics[width=0.95\linewidth]{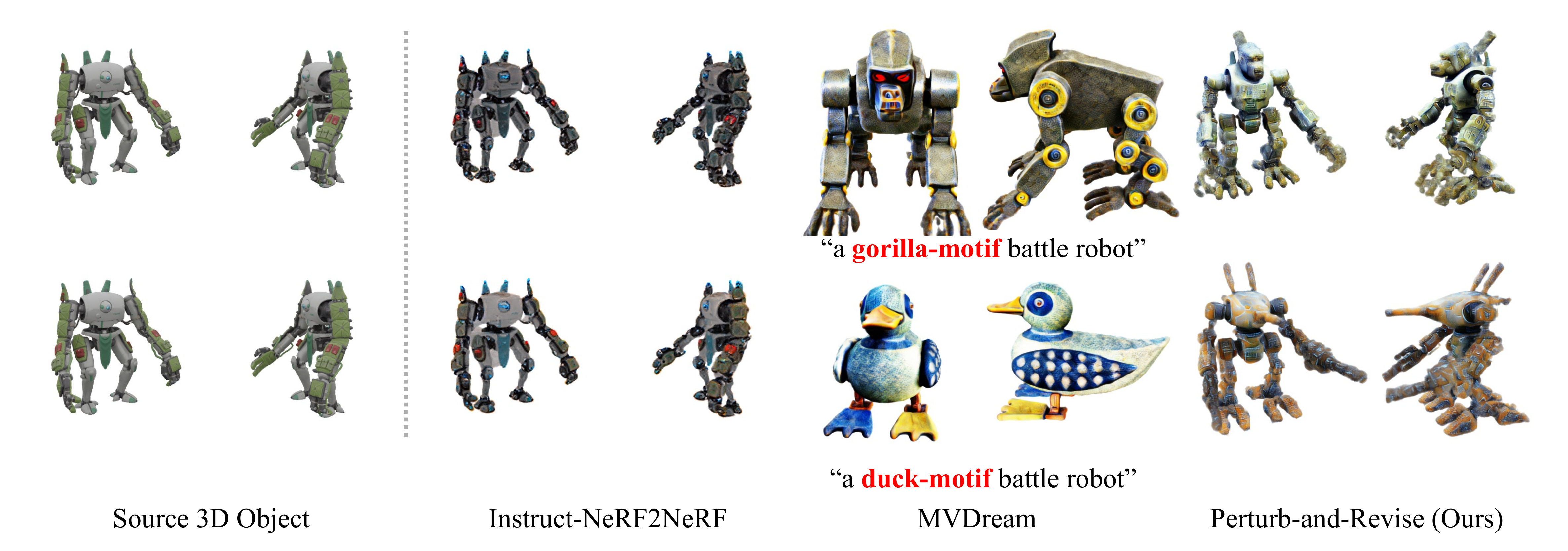}
\vspace{-15pt}
\caption{Comparisons with Instruct-NeRF2NeRF (dataset update)~\cite{haque2023instruct} and MVDream (regeneration)~\cite{shi2023mvdream} in editing Objaverse objects.}
\label{fig:objaverse-2}
\vspace{-5pt}
\end{figure*}

\begin{figure*}[t]
\centering
\begin{subfigure}[]{0.38\textwidth}
\centering
\includegraphics[width=1.0\linewidth]{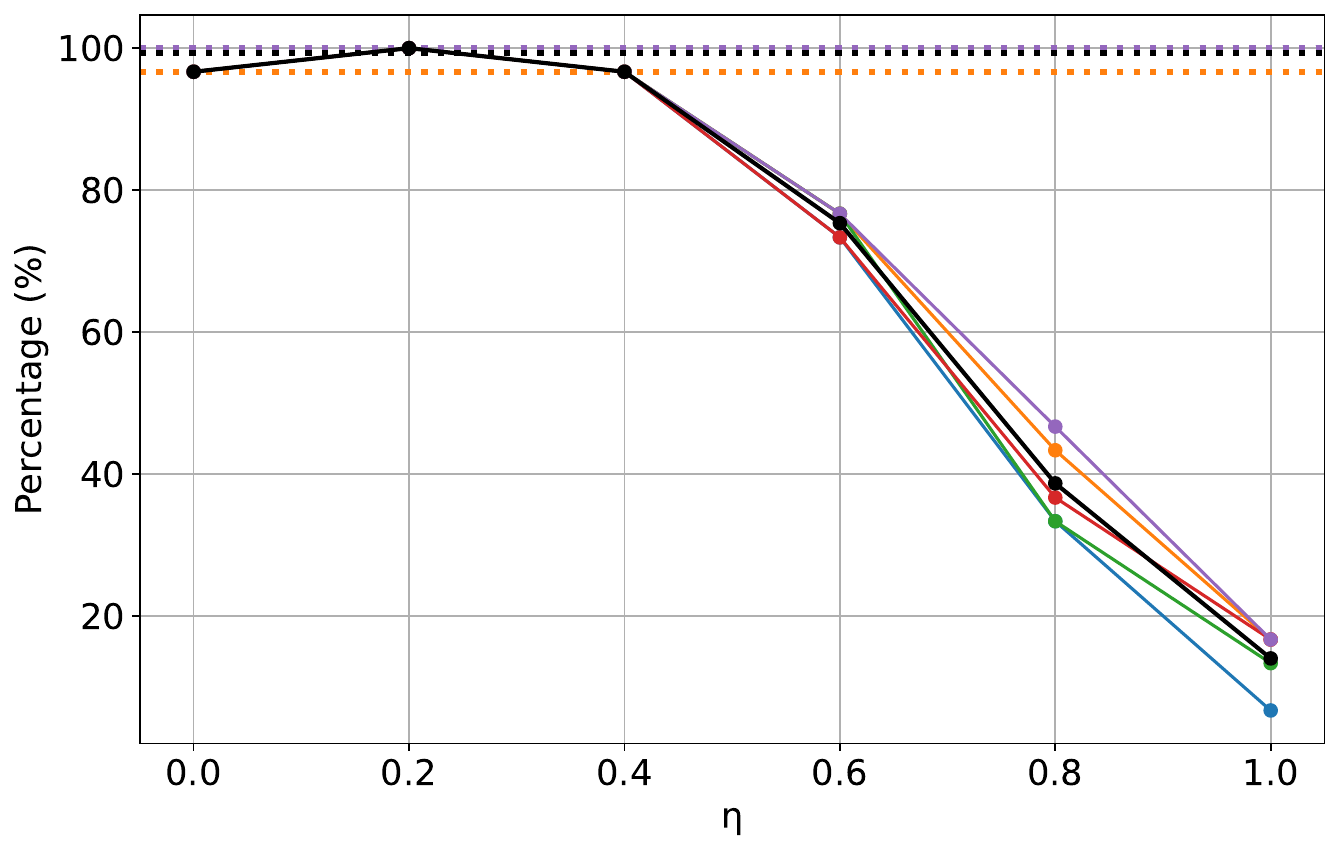}\vspace{-5pt}
\caption{}
\label{fig:abl-p-num-exps}
\end{subfigure}
\begin{subfigure}[]{0.54\textwidth}
\centering
\includegraphics[width=1.0\linewidth]{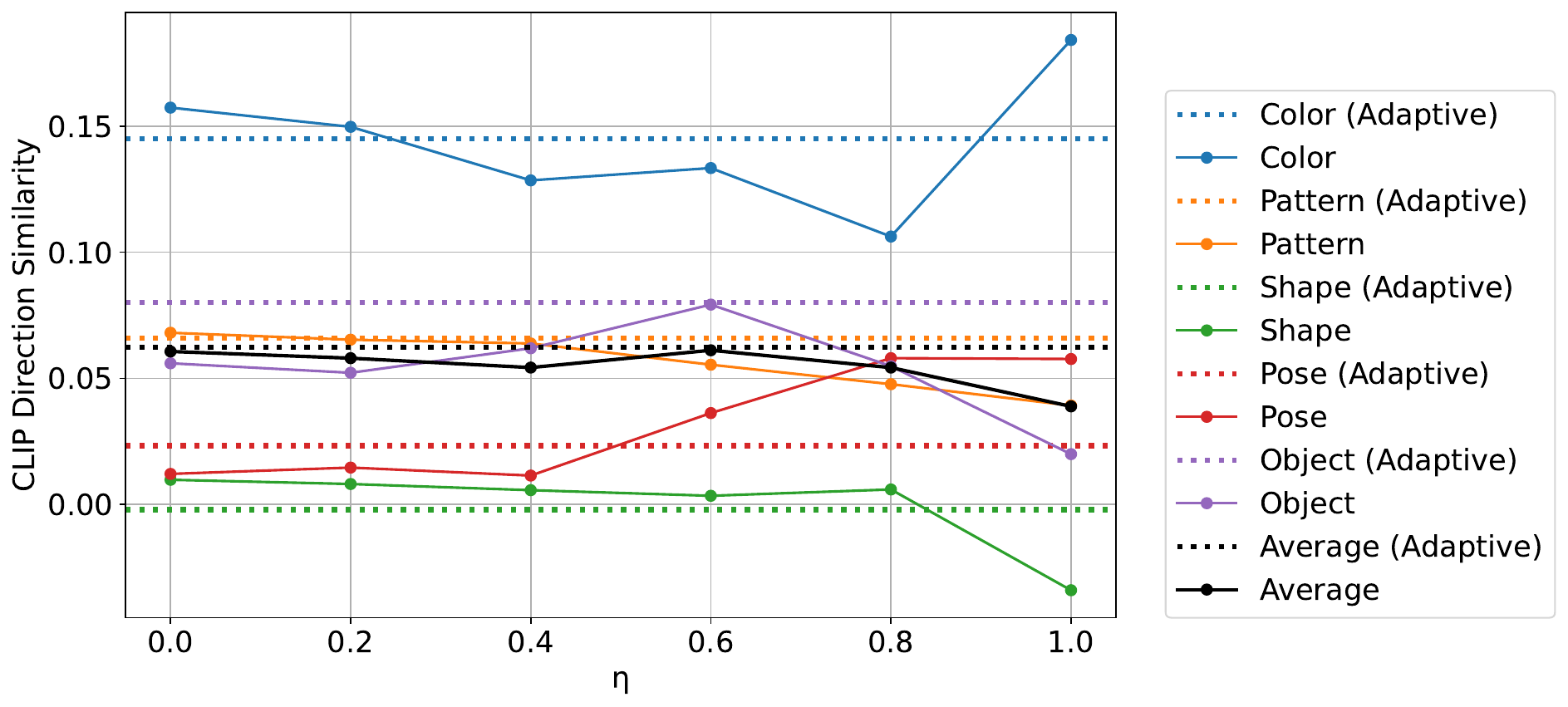}\vspace{-5pt}
\caption{}
\label{fig:abl-p-dir-sim}
\end{subfigure}
\vspace{-12pt}
\caption{Ablation study on the selection of $\eta$. (a) and (b) show the CLIP direction similarity and the percentage of successful experiments (without errors) for different $\eta$ values, respectively. When averaged across all types of edits, our adaptive method achieves near-maximum performance on these metrics compared to all fixed $\eta$ values.}
\label{fig:abl-p}
\vspace{-15pt}
\end{figure*}

\section{Experiments}
\label{sec:experiments}
As baselines, we adopt two distillation-based methods: Score Distillation Sampling (SDS)~\cite{poole2022dreamfusion} and Posterior Distillation Sampling (PDS)~\cite{koo2024posterior}. Bolded red in the prompts indicates the intended edits. We use the exact weighting factors for the losses as originally proposed in \cite{poole2022dreamfusion,koo2024posterior}. For fair comparison, we use MVDream~\cite{shi2023mvdream} as their backbones. Also, we use Instruct-NeRF2NeRF~\cite{haque2023instruct} as our baseline for the iterative dataset update strategy. We use several backbones for each metric and note them in the subscripts, while averaged refers to the average value across all backbones.

\subsection{3D Object Editing}

\paragraph{Fashion object editing.}

In our fashion object editing experiment, we use MVDream~\cite{shi2023mvdream} to generate a synthetic dataset of 3D fashion objects and perform edits using our framework. Specifically, we categorize the fashion object edits into color, pattern, shape, pose, and object edits, and synthesize a total of 150 editing examples, with 30 for each type of edit. In this experiment, our framework takes as input a source object in NeRF with the desired edits to its characteristics, such as color, pattern, shape, pose, and added objects. Note that, unlike some previous work~\cite{koo2024posterior}, we do not require a description of the original object.

The qualitative results, as shown in Figs.~\ref{fig:fashion-timestep} and \ref{fig:fashion-comparison}, demonstrate the versatility introduced by parameter perturbation and our method's capability to perform various types of edits, respectively. In Fig.~\ref{fig:fashion-comparison}, we apply five types of edits: color, pattern, shape, pose, and object additions. We show that our framework supports a wide range of edits, including those requiring pose changes and introducing objects, corroborating its capability for significant geometric modifications, which have rarely been addressed in previous work~\cite{haque2023instruct,koo2024posterior}.

\vspace{-10pt}
\paragraph{General object editing.}
We show a general object editing capability for our method, with objects from Objaverse~\cite{deitke2023objaverse}. We first optimize InstantNGP~\cite{muller2022instant} with $200$ rendered images and camera transforms extracted from an object. Then, we apply our editing method.

We display the qualitative results of our method in Fig.~\ref{fig:objaverse-1}. In addition to textural changes, our method successfully enables structural changes. In Fig.~\ref{fig:objaverse-2}, we also demonstrate our method's ability to perform creative edits anchored by the source 3D object. Here, to demonstrate flexibility, we use two prompts with different animal motifs. We can see that regenerating the object with MVDream produces results that appear completely different from the source 3D object. In contrast, our method successfully anchors the original shape while changing it to align with the edit prompt.

\begin{table}[t]
\footnotesize
\centering
\begin{tabular}{l|ccc}
\toprule
Method & \begin{tabular}[c]{@{}c@{}}CLIP-Dir-\\Sim↑\end{tabular} & \begin{tabular}[c]{@{}c@{}}CLIP-Dir-\\Con↑\end{tabular} & LPIPS↓ \\
\midrule
w/o Refinement & \textbf{0.0624} & 0.7572 & 0.1147 \\
w/ Refinement & 0.0565 & \textbf{0.7642} & \textbf{0.1047} \\
\bottomrule
\end{tabular}
\vspace{-7pt}
\caption{Ablation study on the IPG steps. We present average values across all edit types, prompts, and evaluation backbones. Our approach yields a substantial reduction in $\text{LPIPS}$ while balancing faithfulness to the source object and edit prompt.}
\label{tab:abl-ipg}
\vspace{-15pt}
\end{table}

\subsection{Comparison with Baselines}

We compare our method against baselines to demonstrate our clear advantage in complex edits, as shown in Figs. \ref{fig:fashion-comparison} and \ref{fig:objaverse-1}. For the comparison, we display editing results using SDS with MVDream, Instruct-NeRF2NeRF, and PDS. Additionally, we compare our strategy with dataset update and regeneration strategies in Fig. \ref{fig:objaverse-2}.

Different from other methods, SDS exhibits blurry textures and lacks the capability to perform complex edits. Although it can handle simple changes, it inadvertently changes the appearance and texture of the source objects and cannot perform edits that need extensive geometric changes, as seen in the 3rd, 4th, and 5th rows of Fig.~\ref{fig:fashion-comparison}. While Instruct-NeRF2NeRF successfully achieves simple color edits or symmetric edits, we observe that even when the editing process converges, it is not able to address geometric changes in the objects, such as a change in the pose. PDS, while maintaining similarity to the source object, is unable to make proper edits according to the prompt and cannot deviate far from local minima. As shown in Figs.~\ref{fig:fashion-comparison},~\ref{fig:objaverse-1}, and~\ref{fig:objaverse-2}, our method is capable of various types of edits, including those involving large geometric changes, achieving state-of-the-art results for editing color, appearance, and geometry in a flexible and consistent manner.

We have to consider both the similarity to the source object and faithfulness to the edit prompt. To this end, we evaluate these aspects using CLIP directional similarity~\cite{haque2023instruct} across ViT-B/32, ViT-B/16, and ViT-L/14~\cite{radford2021learning} for measuring faithfulness to the edit prompts and LPIPS~\cite{zhang2018unreasonable} across VGG~\cite{simonyan2014very} and AlexNet~\cite{krizhevsky2017imagenet} for evaluating the similarity between original and edited objects. The results are shown in Table~\ref{tab:quantitative-comparison}. We can see that our method excels at both metrics and achieves a better trade-off than other baselines. This, along with the quantitative evaluation, corroborates the effectiveness of our approach.

\begin{figure}[t]
\centering
\includegraphics[width=1.0\linewidth]{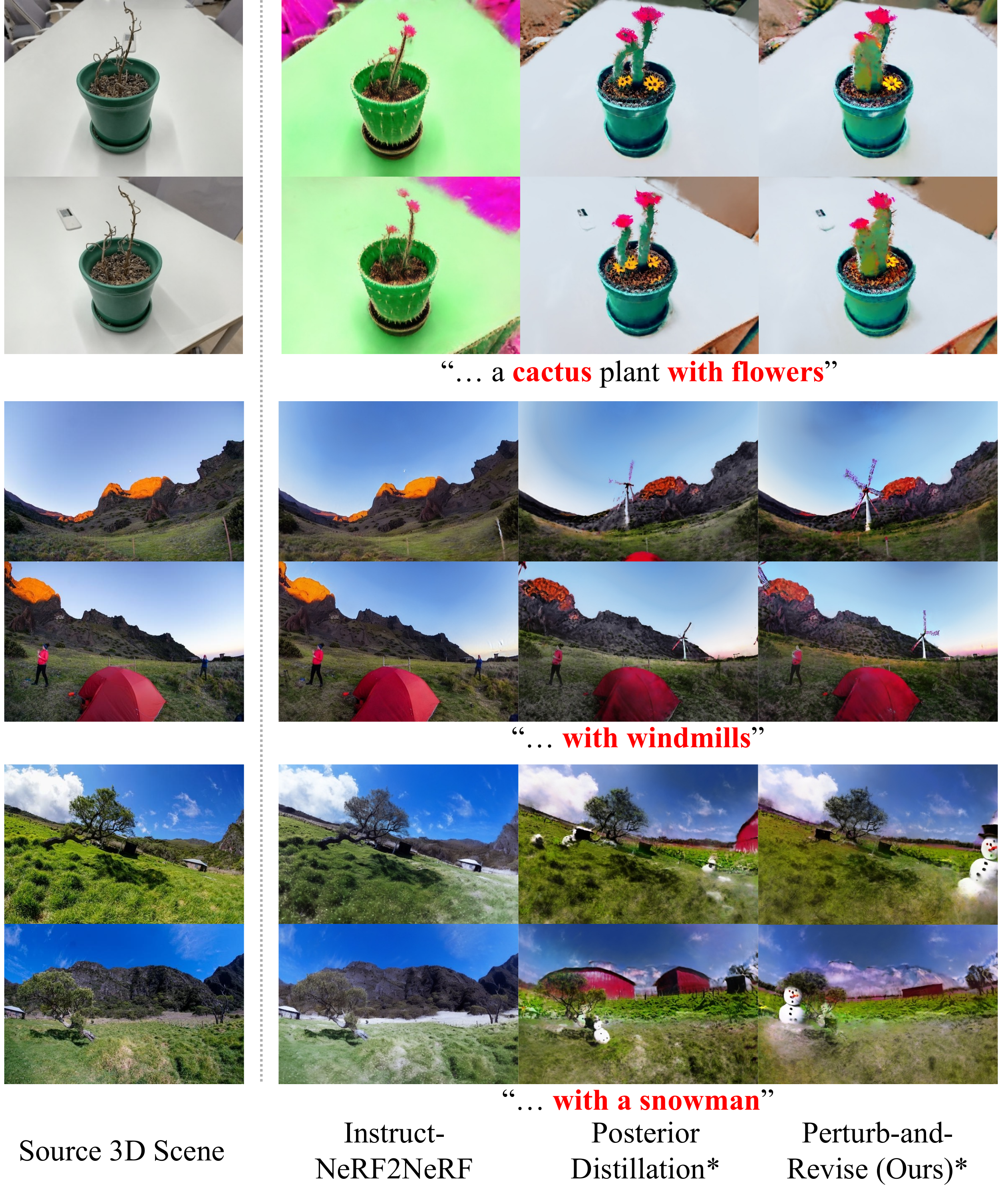}
\vspace{-25pt}
\caption{Real scene editing results. An asterisk (*) indicates we use the exact same update rule and schedule.}
\label{fig:scene}
\vspace{-15pt}
\end{figure}

\subsection{Real Scene Editing}
Our parameter perturbation approach described in Sec.~\ref{sec:parameter-perturbation} can be readily extended to real scene editing scenarios~\cite{haque2023instruct,koo2024posterior}. We present real scene editing results in Figs.~\ref{fig:teaser-bear} and \ref{fig:scene}, building upon Posterior Distillation~\cite{koo2024posterior}. Using the same number of iterations, we can better modify the scene geometry through parameter perturbation. We refer the readers to the supplementary material for details.

\subsection{Ablation Study}

\paragraph{Parameter perturbation.}
In Fig.~\ref{fig:fashion-timestep}, we display the rendered images based on the amount of parameter perturbation and optimization steps. This demonstrates that parameter perturbation facilitates easy alteration of the source object's structure while keeping crucial parts unchanged. Additionally, we present a quantitative study of $\eta$ selection in Fig.~\ref{fig:abl-p}, where we categorize edits into five categories: color, pattern, shape, pose, and object edits. We observe that pose changes and object additions generally require a larger degree of versatility and thus need a larger value of $\eta$ to achieve maximum performance. Furthermore, we demonstrate that, when averaged across all types of edits, the adaptive $\eta$ selection approach outperforms strategies using fixed $\eta \in \{0.0, 0.2, 0.4, 0.6, 0.8\}$ while maintaining high experimental success rates across various edit categories and significantly reducing computational costs compared to grid search. This indicates that our method dynamically selects an effective $\eta$ value while remaining robust in terms of optimization. We present additional metrics in the supplementary material.

\vspace{-10pt}
\paragraph{Identity-preserving gradients.}
We demonstrate the effect of identity-preserving gradients. In Table~\ref{tab:abl-ipg}, we present diverse metrics. Our results show that the refinement steps with IPG achieve a significant decrease in LPIPS while decreasing CLIP directional similarity, due to their trade-off relationship. In addition, we observe an overall increase in CLIP directional consistency~\cite{haque2023instruct}.

\section{Conclusion}

In this paper, we present Perturb-and-Revise (PnR), a framework for text-guided 3D object editing. By introducing adaptive parameter perturbation and identity-preserving gradients, our method enables extensive geometric and appearance changes that adhere to the text prompt while also maintaining fidelity to source objects. Our experiments corroborate that PnR achieves state-of-the-art results across diverse editing tasks without requiring model retraining or multiple input images. We discuss various limitations of our method in the supplementary material for future research.

\vspace{-10pt}
\paragraph{Acknowledgments.}
This work was supported by the UW Reality Lab and Google.

{
    \small
    \bibliographystyle{ieeenat_fullname}
    \bibliography{main}
}

\clearpage
\appendix

\onecolumn
\begin{center}
\Large
\textbf{\thetitle}\\
\vspace{0.5em}Supplementary Material \\
\vspace{1.0em}
\end{center}

\section{Score Distillation as Particle-Based Variational Inference}

Our parameter perturbation and identity gradients build upon the mathematical intuition of the variational score distillation (VSD) approach \cite{wang2024prolificdreamer}, an extension of score distillation sampling (SDS)~\cite{poole2022dreamfusion}. In this context, the parameters of NeRF during distillation are treated as particles.

VSD minimizes the KL divergence between a variational distribution $q^\gamma(x|c)$, which is implicitly modeled by $\gamma$, and the target distribution $p_\phi(x|c)$, which is implicitly modeled by the diffusion model $\phi$. Incorporating timesteps and camera poses, the objective is formulated as follows:
\begin{equation}
    \gamma^*:= \underset{\gamma}{\mathrm{arg}\min}\,\mathbb{E}_{t,\psi}\Big[\frac{\sigma_t}{\alpha_t}w(t)D_\mathrm{KL}(q_t^\gamma(x_t|c, t) \| p_\phi(x_t|c,t))\Big]
\end{equation}

where $\frac{\sigma_t}{\alpha_t}$ and $w(t)$ are diffusion-related weighting factors, and $q_t^\gamma(x_t|c, t)$ and $p_\phi(x_t|c,t)$ represent the distributions of noisy images to be modeled by diffusion models.

To minimize this objective, VSD employs particle-based variational inference based on Wasserstein gradient flow, as detailed in \cite{chen2018unified,liua2022geometry,wang2019function,dong2022particle}. Specifically, the Wasserstein gradient flow satisfies:
\begin{equation}
\frac{\partial \gamma_\tau}{\partial \tau} = \nabla \cdot (\gamma_\tau\nabla(\frac{\partial E}{\partial \gamma_\tau}(\gamma_\tau)))
\end{equation}
In our case, the energy functional $E$ is defined as follows:
\begin{equation}
    E(\gamma) := \mathbb{E}_{t, \psi}\Big[\frac{\sigma_t}{\alpha_t}w(t)D_\mathrm{KL}(q_t^\gamma(x_t|c, t) \| p_\phi(x_t|c,t))\Big]
\label{eq:energy}
\end{equation}

In the particle-based variational inference, particles represent samples from the variational distribution. A set of $M$ particles $\{\theta^{(i)}\}_{i=1}^M \sim \gamma$ is iteratively updated following the velocity of particles~\cite{chen2018unified}: $\frac{d\theta_\tau}{d\tau} = \nabla(\frac{\partial E}{\partial \gamma_\tau}(\gamma_\tau))$. With the energy function in Eq.~\ref{eq:energy}, the particles follow the ordinary differential equation (ODE):
\begin{align}
    \frac{d\theta_\tau}{d\tau}=&-\mathbb{E}_{t,\epsilon,\psi}\Big[w(t)\big(-\sigma_t\nabla_{x_t}\log p_\phi(x_t|c,t) - (-\sigma_t\nabla_{x_t}\log q_t^{\gamma_\tau}(x_t|c, t))\frac{\partial g(\theta_\tau,\psi)}{\partial \theta_\tau}\big)\Big]
\end{align}
where $\tau$ denotes the ODE time, constrained to $\tau\geq0$, and $\gamma_\tau$ progressively evolves toward the optimal distribution $\gamma^*$ as $\tau\to\infty$. In this VSD framework, the gradient of the SDS loss is a specific instance of the equation~\cite{wang2024prolificdreamer}, where a single particle represents the entire distribution.

\clearpage

\section{Resulting Distribution from Parameter Interpolation (Sec. 4.1)}

Here, we show that interpolating parameters with $\eta \in [0, 1]$ results in a versatile sampling distribution that interpolates between a point mass at $\theta_\textrm{src}$ and the initial distribution. Given a source parameter $\theta_{\textrm{src}}$ and an initial distribution $\mathcal{P}(\Theta_0)$ with bounded variance $\sigma^2$, we define the parameter perturbation as:
\begin{align}
\theta_{\textrm{perturbed}} &= (1-\eta)\theta_{\textrm{src}} + \eta\theta_0, \quad \theta_0 \sim \mathcal{P}(\Theta_0), \quad \eta \in [0,1]
\end{align}
Using the change of variables formula with transformation $T(\theta_0) = (1-\eta)\theta_{\textrm{src}} + \eta\theta_0$ and its inverse $T^{-1}(\theta_{\textrm{perturbed}}) = (\theta_{\textrm{perturbed}} - (1-\eta)\theta_{\textrm{src}})/\eta$:
\begin{align}
p(\theta_{\textrm{perturbed}}) = \mathcal{P}(\Theta_0)(T^{-1}(\theta_{\textrm{perturbed}})) \cdot |\det(J_{T^{-1}})|
\end{align}
Since the Jacobian matrix is $J_{T^{-1}} = \frac{1}{\eta}I_d$, where $I_d$ is the $d$-dimensional identity matrix, we have:
\begin{align}
p(\theta_{\textrm{perturbed}}) &= \frac{1}{\eta^d} \mathcal{P}(\Theta_0)\left(\frac{\theta_{\textrm{perturbed}} - (1-\eta)\theta_{\textrm{src}}}{\eta}\right)
\end{align}
Here, $\eta$ controls the degree of interpolation through both a scale factor $\frac{1}{\eta^d}$ and the argument $(\theta_{\textrm{perturbed}} - (1-\eta)\theta_{\textrm{src}})/\eta$ of $\mathcal{P}(\Theta_0)$.

For $\eta \to 1$, both terms approach simple limits:
\begin{align}
\lim_{\eta \to 1} p(\theta_{\textrm{perturbed}}) &= \lim_{\eta \to 1} \frac{1}{\eta^d} \mathcal{P}(\Theta_0)\left(\frac{\theta_{\textrm{perturbed}} - (1-\eta)\theta_{\textrm{src}}}{\eta}\right) \\
&= \mathcal{P}(\Theta_0)(\theta_{\textrm{perturbed}})
\end{align}

For $\eta \to 0$, we consider the distribution of $\theta_{\textrm{perturbed}}$. By Chebyshev's inequality, for any $\varepsilon > 0$:
\begin{align}
P(|\theta_{\textrm{perturbed}} - \mathbb{E}[\theta_{\textrm{perturbed}}]| \geq \varepsilon) &\leq \frac{\eta^2\sigma^2}{\varepsilon^2} \to 0 \quad \text{as} \quad \eta \to 0
\end{align}
Moreover, since $\mathbb{E}[\theta_{\textrm{perturbed}}] \to \theta_{\textrm{src}}$ as $\eta \to 0$:
\begin{align}
P(|\theta_{\textrm{perturbed}} - \theta_{\textrm{src}}| \geq \varepsilon) &\to 0 \quad \text{as} \quad \eta \to 0
\end{align}
This proves convergence in probability to $\theta_{\textrm{src}}$. The $\frac{1}{\eta^d}$ factor ensures that the total probability remains 1, while the concentration around $\theta_{\textrm{src}}$ becomes arbitrarily tight as $\eta \to 0$, characterizing convergence to:
\begin{align}
\lim_{\eta \to 0} p(\theta_{\textrm{perturbed}}) = \delta(\theta_{\textrm{perturbed}} - \theta_{\textrm{src}})
\end{align}

Thus, we have shown that the interpolation of parameters results in an interpolation between two extremes: a point mass at $\theta_\textrm{src}$ and the initial distribution, and the parameter $\eta$ controls the degree of interpolation, i.e., the versatility.

\clearpage

\vspace*{\fill}

\begin{algorithm}[h]
\SetAlgoLined
\DontPrintSemicolon
\SetKwFunction{FMixEdit}{ParameterPerturbation}
\SetKwProg{Fn}{Function}{:}{}
\Fn{\FMixEdit{$\eta$}}{
    $\theta_\text{new} \leftarrow$ Initialize new geometry instance\;
    
    \For{$(\theta_c, \theta_n, \theta_i)$ \textbf{in} zip($\theta_\text{current}$, $\theta_\text{new}$, $\theta_\text{init}$)}{
        $\theta_c \leftarrow (1-\eta)\theta_i + \eta\theta_n$ \tcp*{Parameter interpolation}
    }
    Free memory and clear cache\;
}
\caption{Parameter Perturbation}
\label{alg:parameter-perturbation}
\end{algorithm}

\begin{algorithm}[h]
\SetAlgoLined
\DontPrintSemicolon
\KwIn{Empty loss history list $\mathcal{L}$, minimum loss decrease $\Delta_\text{min}$, maximum parameter perturbation $\eta_\text{max}$}
\KwIn{Initial NeRF parameters $\theta_\text{init}$}
\SetKwFunction{FTrainingStep}{TrainingStep}
\SetKwFunction{FCalcDecrease}{LossDecrease}
\SetKwFunction{FMixEdit}{ParameterPerturbation}
\SetKwFunction{FDetermineP}{DetermineEta}
\SetKwProg{Fn}{Function}{:}{}
\Fn{\FTrainingStep}{
    \If{$|\mathcal{L}| = 50$}{
        $\Delta\mathcal{L} \leftarrow $ \FCalcDecrease{$\mathcal{L}$}\;
        $\eta \leftarrow $ \FDetermineP{$\Delta\mathcal{L}$, $\Delta_\text{min}$, $\eta_\text{max}$}\;
        \FMixEdit{$\eta$}\;
    }
    Proceed with training step\;
    $\mathcal{L} \leftarrow \mathcal{L} \oplus \{\text{Current step training loss}\}$\;
}
\Fn{\FCalcDecrease{$\mathcal{L}$}}{
    $\mathcal{L}_\text{final} \leftarrow \frac{1}{10}\sum_{i=|\mathcal{L}|-10}^{|\mathcal{L}|} \mathcal{L}_i$ \tcp*{Average of last 10 losses}
    $\mathcal{L}_\text{init} \leftarrow \frac{1}{10}\sum_{i=1}^{10} \mathcal{L}_i$ \tcp*{Average of first 10 losses}
    \Return{$\mathcal{L}_\text{final} - \mathcal{L}_\text{init}$}
}
\Fn{\FDetermineP{$\Delta\mathcal{L}$, $\Delta_\text{min}$, $\eta_\text{max}$}}{
    \Return{$\max(0, \eta_\text{max}(1 - 2^{-(\Delta\mathcal{L} + \Delta_\text{min})/\Delta_\text{min}}))$}\;
}
\caption{Parameter Perturbation with Adaptive $\eta$ Selection}
\label{alg:eta-determination}
\end{algorithm}

\section{Algorithms for Parameter Perturbation and Adaptive \texorpdfstring{$\eta$}{} Selection}

We present the complete algorithms for parameter perturbation in Alg.~\ref{alg:parameter-perturbation} and the adaptive $\eta$ selection method in Alg.~\ref{alg:eta-determination}. The underlying intuition for the adaptive $\eta$ selection algorithm is that there exists a minimum loss decrease $\Delta_\textrm{min}$ required for parameter perturbation and a maximum parameter perturbation $\eta_\textrm{max}$ that can be applied without resulting in complete regeneration of the object. Here, we have two parameters to control, $\Delta_\textrm{min}$ and $\eta_\textrm{max}$. $\Delta_\textrm{min}$ is set to 1000 based on observations that it achieves near-optimal CLIP directional similarity and CLIP directional consistency, as shown in Table~\ref{tab:abl-delta-min}. $\eta_\textrm{max}$ is set to 0.6 based on the finding that the percentage of successful experiments drops significantly when $\eta$ exceeds 0.6, as shown in the main paper.

\vspace*{\fill}

\clearpage

\section{Additional Analyses}

\paragraph{Intermediate results from real scene editing.}

\begin{figure}[t]
\centering
\begin{subfigure}[b]{1.0\linewidth}
    \centering
    \includegraphics[width=\linewidth]{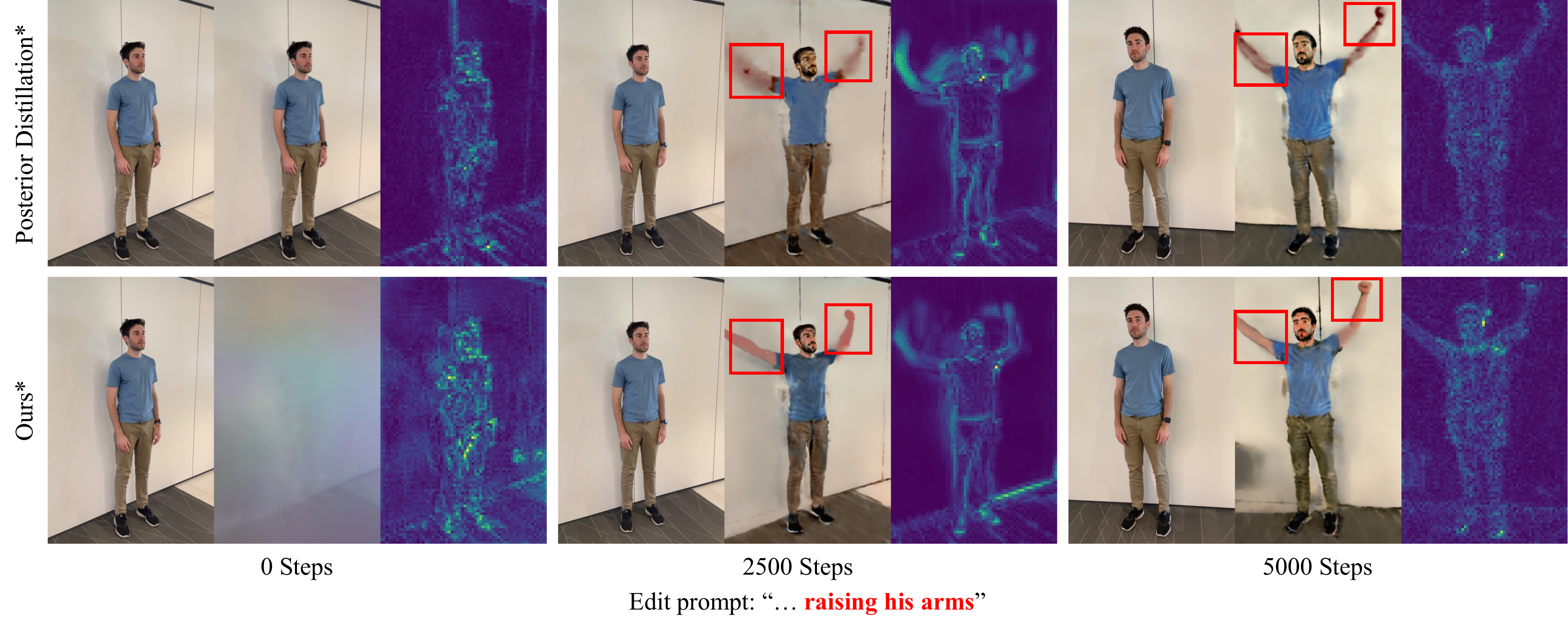}
    \label{fig:intermediate-1}
\end{subfigure}

\begin{subfigure}[b]{1.0\linewidth}
    \centering
    \includegraphics[width=\linewidth]{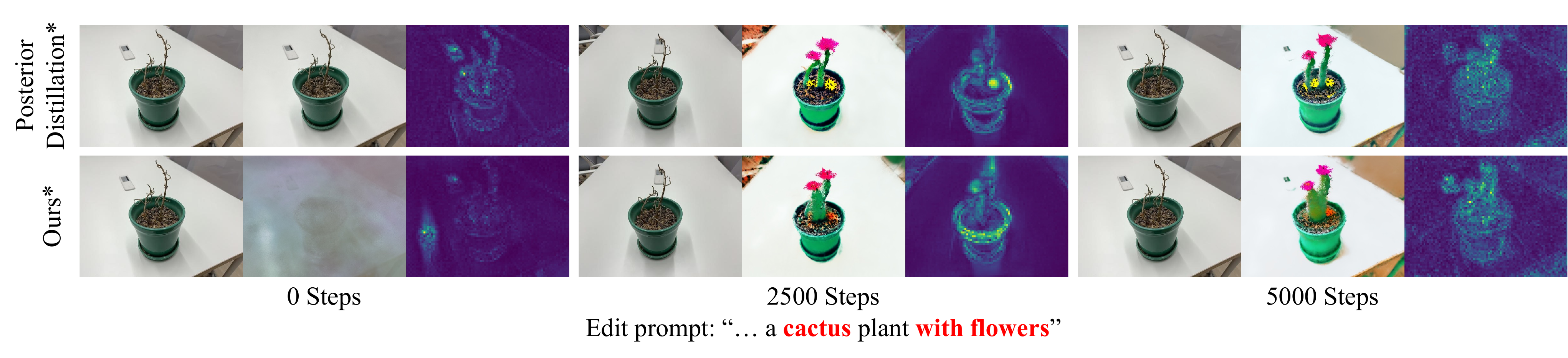}
    \label{fig:intermediate-2}
\end{subfigure}
\caption{Original scene, edited scene, and image-level gradients are shown at 0, 2500, and 5000 optimization steps. We can see that the density forms earlier and changes drastically even when the perturbation is large and barely has any structure.}
\label{fig:intermediate}
\end{figure}

In Fig.~\ref{fig:intermediate}, we show intermediate results from a real scene editing experiment. In this experiment, we aim to make the person raise the arms. Despite the initialization having little 3D structure, it converges faster with the same number of optimization steps. We can see that the density near the raised arms quickly converges with parameter perturbation, while the original PDS~\cite{koo2024posterior} generates blurry results. This demonstrates the effectiveness of our parameter perturbation approach in various editing scenarios.

\paragraph{Additional ablation study.}

\begin{figure}
\centering
\begin{subfigure}[]{0.38\textwidth}
\centering
\includegraphics[width=1.0\linewidth]{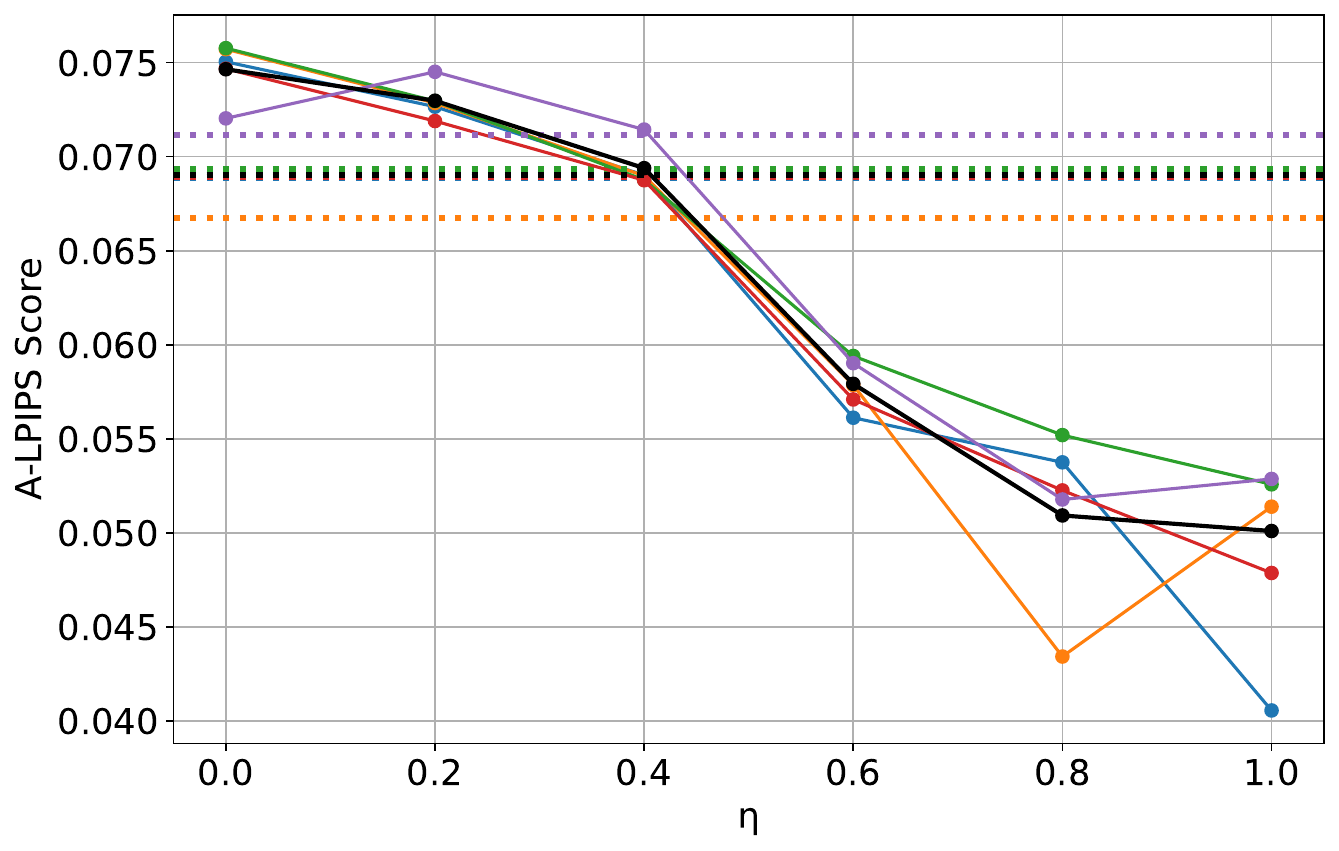}
\caption{}
\label{fig:abl-p-a-lpips}
\end{subfigure}
\begin{subfigure}[]{0.54\textwidth}
\centering
\includegraphics[width=1.0\linewidth]{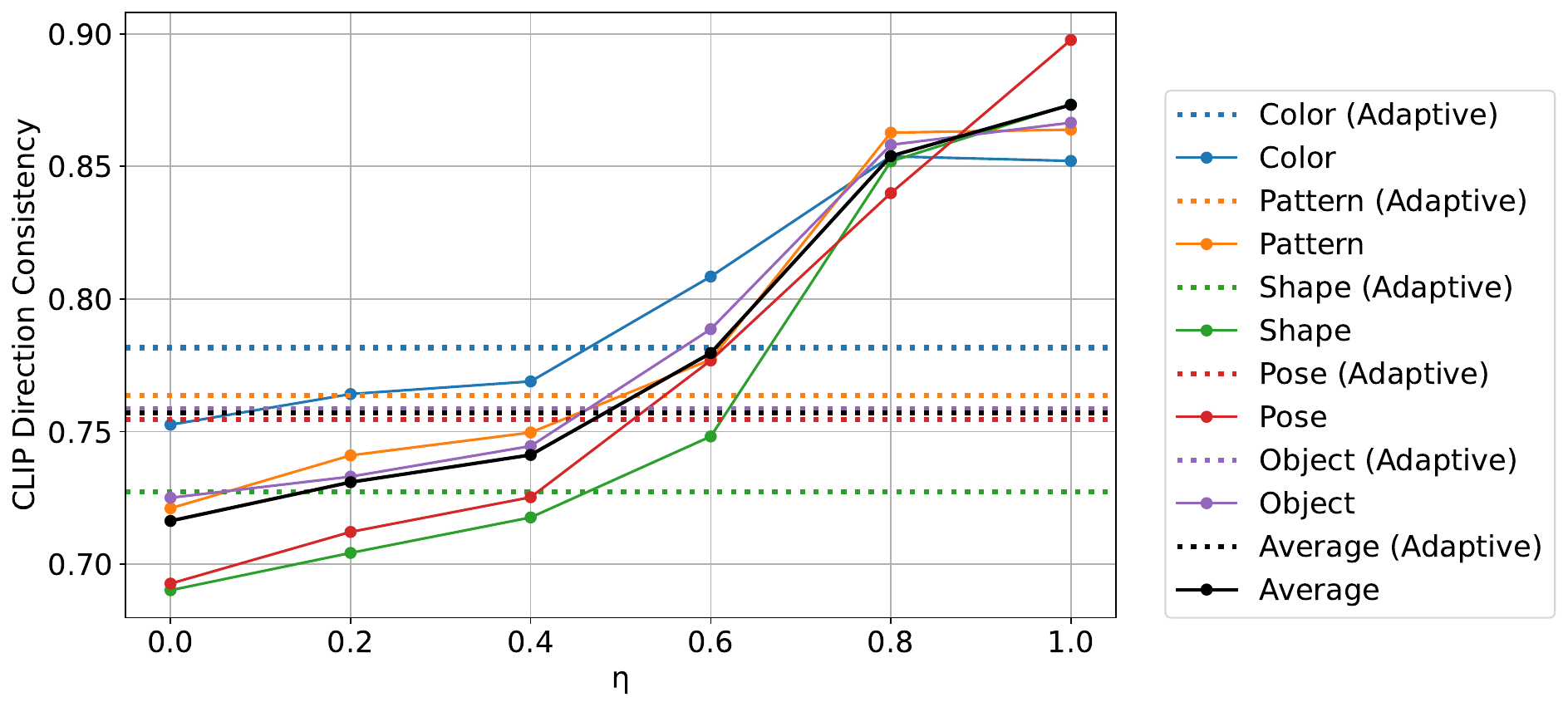}
\caption{}
\label{fig:abl-p-dir-con}
\end{subfigure}
\caption{Additional visualizations for the selection of $\eta$. (a) and (b) show the A-LPIPS (lower is better) and CLIP directional consistency (higher is better) for different $\eta$ values, respectively.}
\label{fig:abl-p-supple}
\end{figure}

We present additional ablation study results on $\Delta_\text{min}$ in Table~\ref{tab:abl-delta-min}. Additionally, we examine the effects of different values for $\lambda_\text{L1}$ and $\lambda_\text{p}$ in Table~\ref{tab:abl-l1-p}. Our results demonstrate that our method is relatively robust to these parameters, with our chosen values achieving a near-optimal balance across metrics. In addition, in Fig.~\ref{fig:abl-p-supple}, we display additional visualizations for the selection of $\eta$. A-LPIPS~\cite{hong2023debiasing} is a metric for view consistency between adjacent frames, and CLIP directional consistency~\cite{haque2023instruct} is a metric that computes how much the editing directions differ across frames. Considering that we showed in the main paper that using fixed values of $\eta \geq 0.6$ had a higher likelihood of causing errors, our method outperforms approaches using fixed values of $\eta < 0.6$ in both metrics while maintaining lower error rates.

\begin{figure}[t]
\centering
\includegraphics[width=0.9\linewidth]{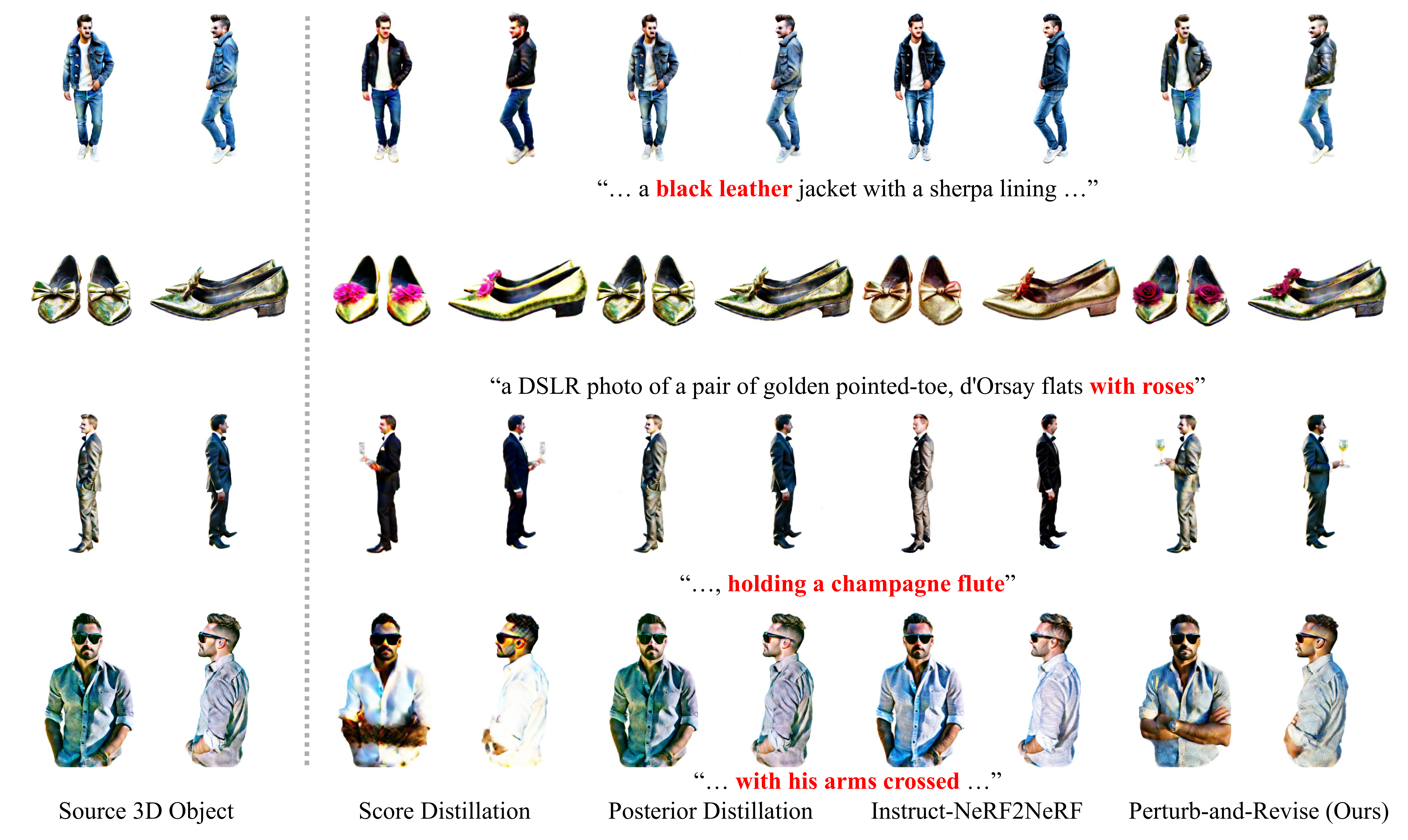}
\vspace{-10pt}
\caption{Additional comparisons of fashion object editing with Score Distillation~\cite{poole2022dreamfusion}, Posterior Distillation~\cite{koo2024posterior}, Instruct-NeRF2NeRF~\cite{haque2023instruct}, and Perturb-and-Revise (ours).}
\label{fig:fashion-2}
\vspace{-10pt}
\end{figure}

\begin{figure}[t]
\centering
\includegraphics[width=0.9\linewidth]{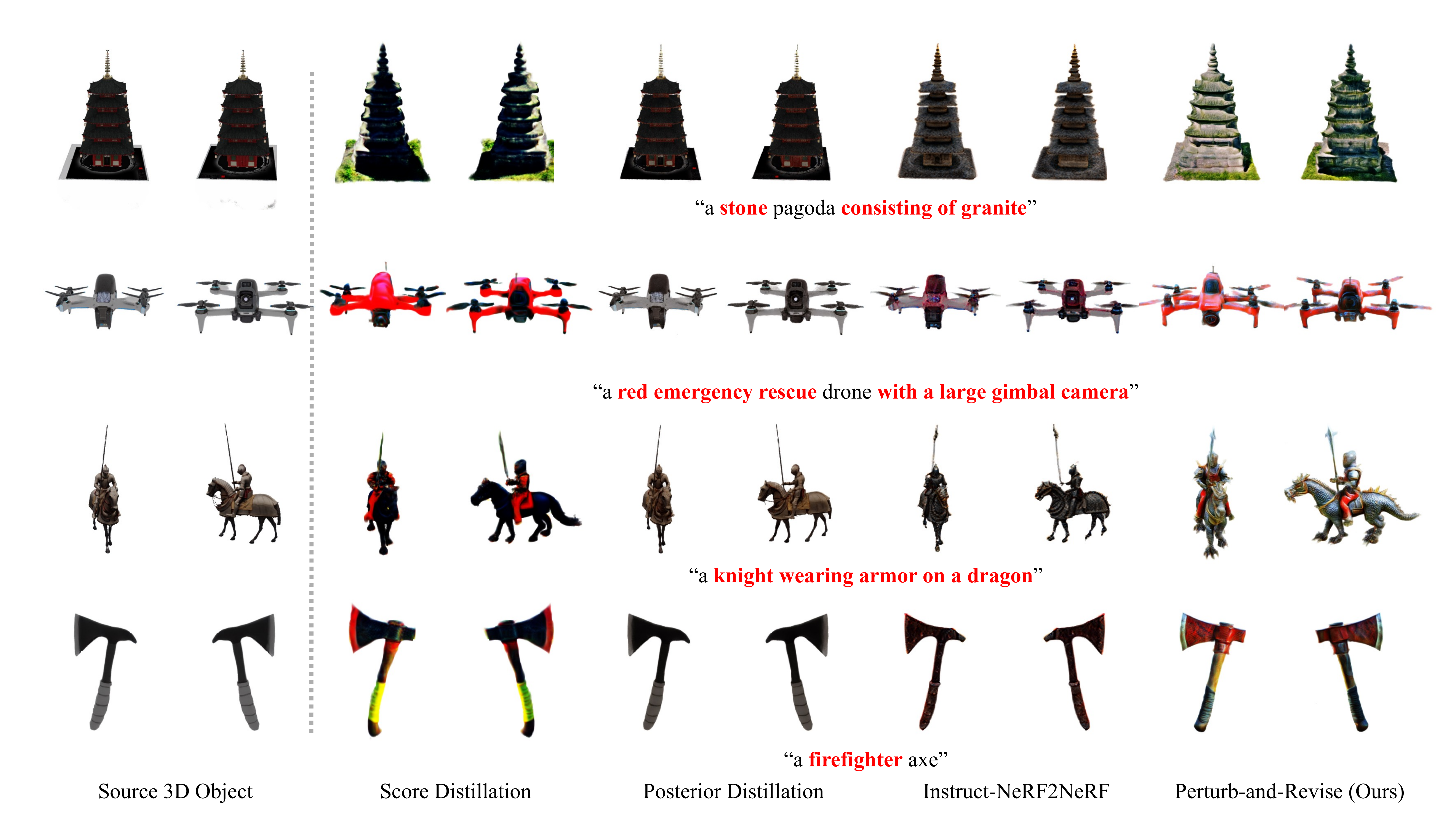}
\vspace{-10pt}
\caption{Additional comparisons of general object editing with Score Distillation~\cite{poole2022dreamfusion}, Posterior Distillation~\cite{koo2024posterior}, Instruct-NeRF2NeRF~\cite{haque2023instruct}, and Perturb-and-Revise (ours).}
\label{fig:objaverse-3}
\vspace{-10pt}
\end{figure}

\paragraph{Additional comparisons.}
In Figs.~\ref{fig:fashion-2} and \ref{fig:objaverse-3}, we showcase additional comparisons with the baseline methods.

\paragraph{Comparisons in 360° views.}
We present qualitative comparisons with 360° views on our project page.

\begin{table}[t]
\small
\centering
\begin{tabular}{l|ccc}
\toprule
Method & CLIP-Dir-Sim$_\text{averaged}$↑ & CLIP-Dir-Con$_\text{averaged}$↑ & LPIPS$_\text{averaged}$↓ \\
\midrule
$\Delta_\text{min}=500$ & 0.061 & 0.757 & 0.112 \\
$\Delta_\text{min}=1000$ & 0.062 & 0.757 & 0.115 \\
$\Delta_\text{min}=2000$ & 0.060 & 0.754 & 0.111 \\
\bottomrule
\end{tabular}
\caption{Experiment controlling $\Delta_\text{min}$.}
\label{tab:abl-delta-min}
\end{table}

\begin{table}[t]
\small
\centering
\begin{tabular}{l|ccc}
\toprule
Method & CLIP-Dir-Sim$_\text{averaged}$↑ & CLIP-Dir-Con$_\text{averaged}$↑ & LPIPS$_\text{averaged}$↓ \\
\midrule
$\lambda_\text{L1} = 10000, \lambda_\text{p} = 100$ & 0.057 & 0.777 & 0.115 \\
$\lambda_\text{L1} = 30000, \lambda_\text{p} = 300$ & 0.057 & 0.764 & 0.105 \\
$\lambda_\text{L1} = 50000, \lambda_\text{p} = 500$ & 0.051 & 0.752 & 0.091 \\
\bottomrule
\end{tabular}
\caption{Experiment controlling $\lambda_\text{L1}$ and $\lambda_\text{p}$.}
\label{tab:abl-l1-p}
\end{table}

\begin{figure}[t]
\centering
\includegraphics[width=0.8\linewidth]{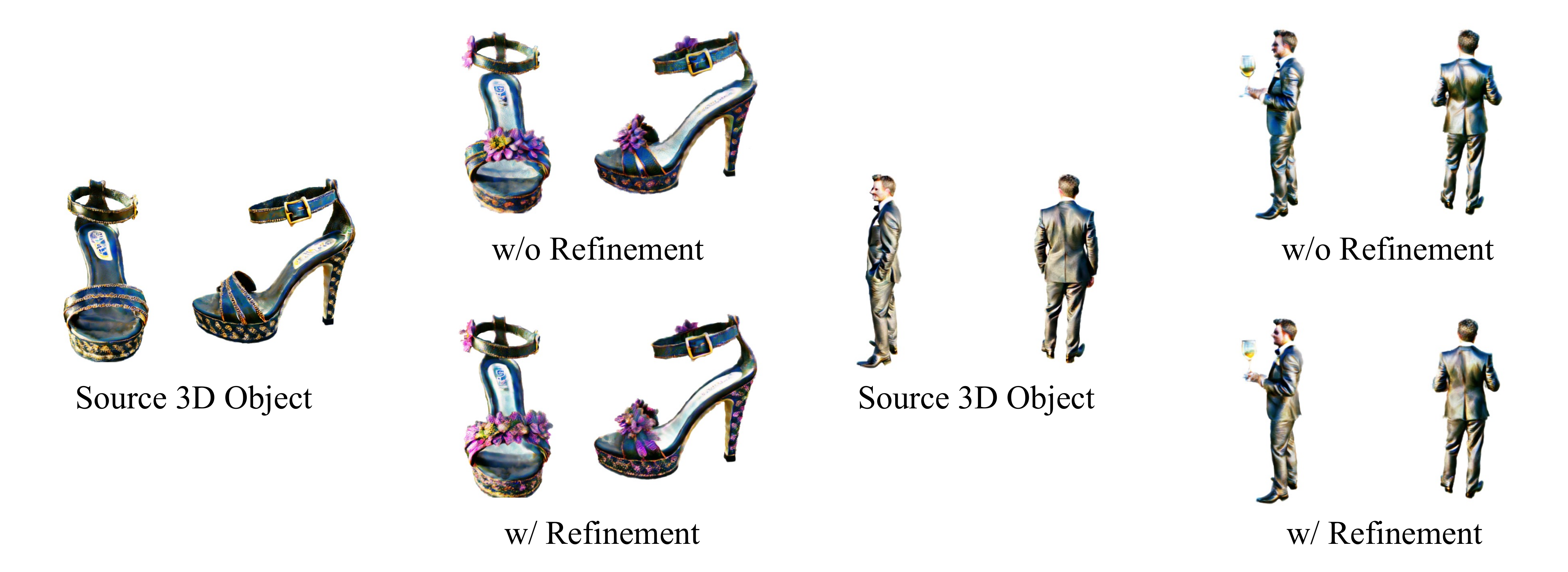}
\vspace{-10pt}
\caption{Effects of IPG refinement steps. IPG refinement steps restore changed attributes that were not explicitly mentioned in the edit prompt during the editing process, for example, the support part of the strap and the subtle details in the color and texture of the tuxedo.}
\label{fig:abl-ipg}
\vspace{-10pt}
\end{figure}

\paragraph{Effect of IPG.}

In Fig.~\ref{fig:abl-ipg}, we demonstrate refinement outcomes through IPG and the generative ODE. A notable IPG attribute is its preservation of areas in the 3D object not explicitly specified for modification within the editing prompt.

\paragraph{Extension to 3DGS.}

\begin{wrapfigure}{r}{0.4\textwidth}
\vspace{-20pt}
\centering
\includegraphics[width=0.35\textwidth]{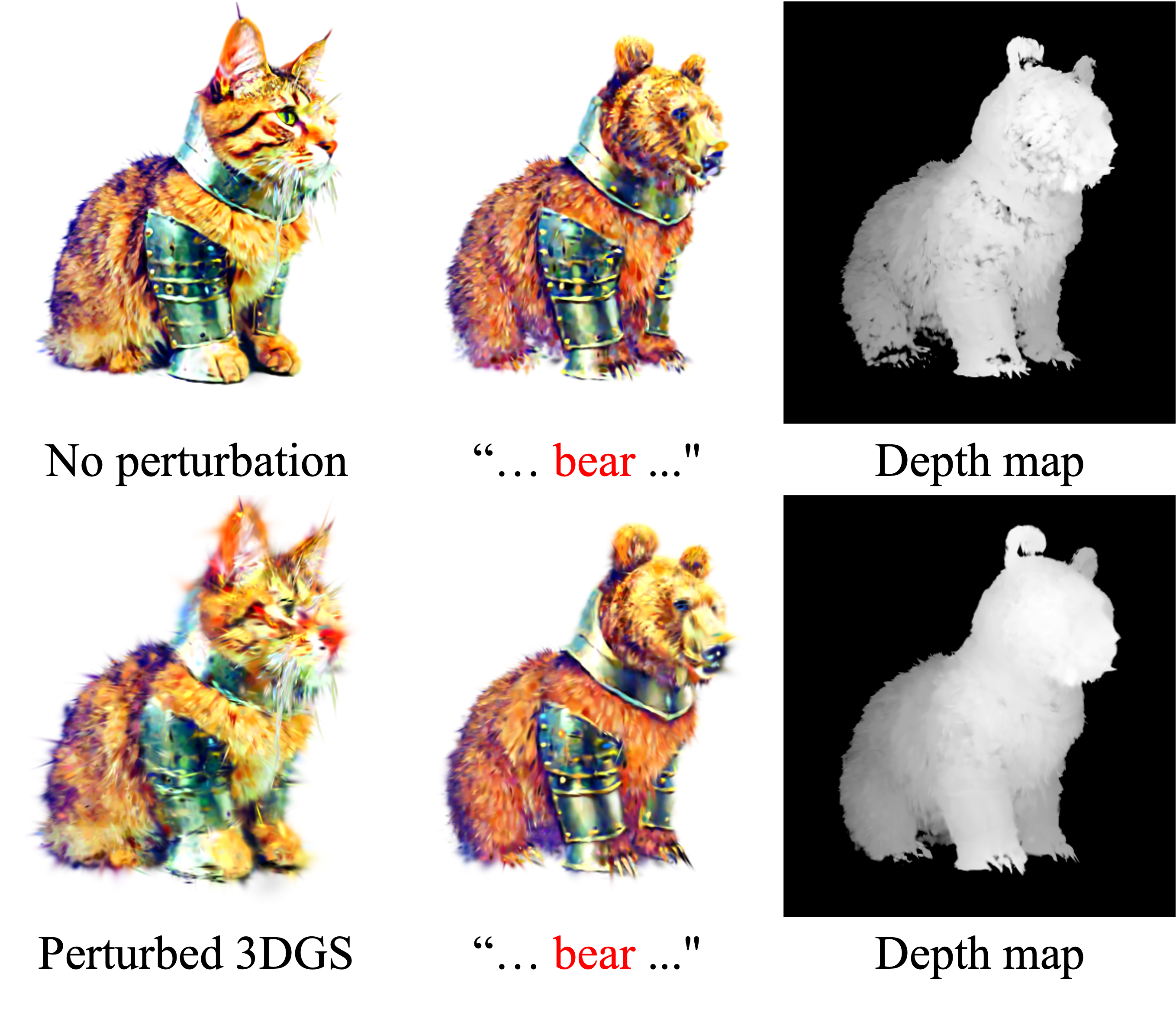}
\caption{3DGS experiment. The perturbation approach improves the depth map and overall geometry of the resulting object.}
\label{fig:3dgs}
\vspace{-40pt}
\end{wrapfigure}

Adapting our method to 3DGS~\cite{kerbl20233d} presents unique challenges, particularly in addressing adaptive densification and the ill-defined interpolation of initial and optimized Gaussians. However, our preliminary experiment, based on 3DGS generated by LucidDreamer~\cite{liang2024luciddreamer}, with mean and variance perturbation shows promising results, yielding improved depth maps and geometry under identical conditions (Fig.~\ref{fig:3dgs}).

\section{Implementation Details}
\label{sec:implementation}

\paragraph{Optimization steps.}
For all perturbation values, we perform 1.5k editing steps, significantly fewer than the 10k steps required for regeneration~\cite{shi2023mvdream}. We set a resolution milestone in fashion object editing at which the rendering resolution changes for efficacy to half the number of editing steps. We perform 1k additional refinement steps, making the total runtime similar to 1.5k steps of PDS and thus highly efficient.

\paragraph{Identity-preserving gradients.}
For the identity-preserving gradients in Sec.~4.3, we adopt a combination of perceptual and L1 losses, finding this more stable and less fragile to noise than using only L1 or L2 loss. Specifically, we choose $\lambda_{\text{L1}}=300.0$ and $\lambda_{\text{p}}=30000.0$, with an annealed schedule, i.e., we linearly decrease them to $0$ until the halfway point of the steps. We present an ablation study on the scales of $\lambda_{\text{p}}$ and $\lambda_{\text{L1}}$ in Table~\ref{tab:abl-l1-p}.

\paragraph{Timestep annealing.}
In the original Score Distillation~\cite{poole2022dreamfusion}, Instruct-NeRF2NeRF~\cite{haque2023instruct}, and Posterior Distillation~\cite{koo2024posterior} papers, a fixed schedule, $\Sigma := \mathcal{U}(0.02, 0.98)$, is utilized. Contrary to this fixed schedule, and considering that our editing purpose does not inherently start from random parameters, we adopt a schedule in which $\Sigma(0) = \mathcal{U}(0.75, 0.75)$, a range to be decreased to $(0.02, 0.4)$ by the time $80\%$ of the total editing steps are reached.

\paragraph{NeRF representation.}
Technically, the parameter perturbation method can be applied to arbitrary representations whose parameters are initialized from a distribution and optimized. For computational efficiency while maintaining high quality of 3D objects, we choose InstantNGP~\cite{muller2022instant} as our NeRF implementation.

\paragraph{Real scene editing.}

We show in the main paper that our parameter perturbation approach can be readily extended to real scene editing scenarios~\cite{haque2023instruct}. Using Nerfstudio's implementation of Nerfacto~\cite{tancik2023nerfstudio} as the representation, we integrate Instruct-NeRF2NeRF~\cite{haque2023instruct} without modifications. For this experiment, we build upon the distillation method proposed in PDS~\cite{koo2024posterior}, use Stable Diffusion v1-5~\cite{rombach2022high} as the backbone, and set $\eta = 0.6$. Besides the perturbation, we use the exact same update rule as PDS. We reduce the timesteps by half and omit the selection and refinement steps for both PDS and our method to manage computational complexity and to show only the effect of the parameter perturbation approach. Note that PDS has an internal preservation term~\cite{koo2024posterior}, and we use it as is for scene editing. Even with these shortened iteration steps, our parameter perturbation approach enables extensive geometric editing of the scene.

\clearpage

\begin{figure}[t]
\centering
\includegraphics[width=0.7\linewidth]{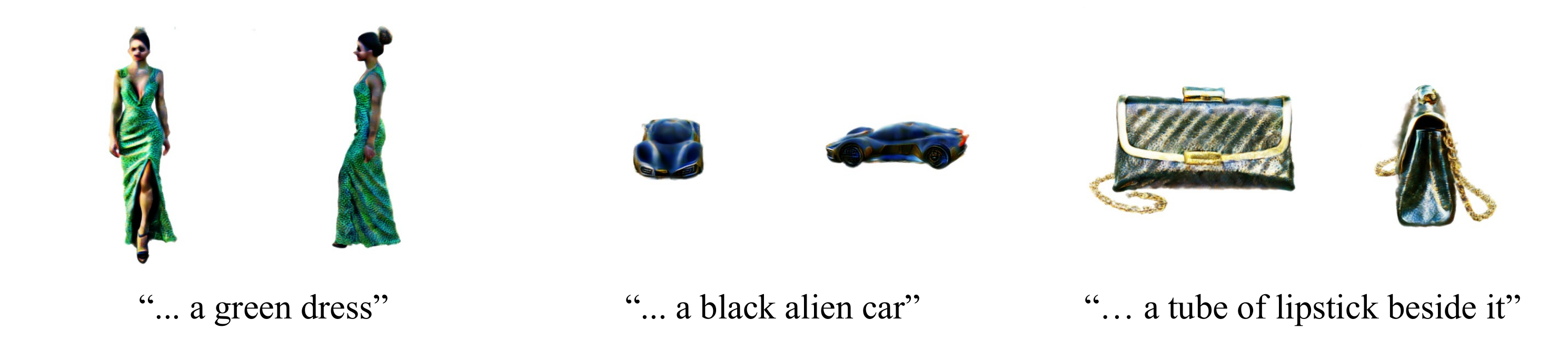}
\vspace{-10pt}
\caption{Failure cases.}
\label{fig:failure}
\vspace{-10pt}
\end{figure}

\section{Computational efficiency}

Our approach requires approximately 7 minutes with IPG and 4 minutes without, to produce meaningful results. This is faster than the 13 minutes needed for Instruct-NeRF2NeRF. We attribute this to Instruct-NeRF2NeRF requiring updates to the entire dataset, while our method of parameter perturbation and timestep annealing benignly affects the optimization process of an object. Note that completely regenerating an object requires around 26 minutes.

\section{Limitations}

Although we address limitations of previous work, our method inherits some limitations from pre-trained diffusion models ~\cite{shi2023mvdream,rombach2022high}, such as color biases and saturation artifacts (Fig.~\ref{fig:failure}). While our method can handle pose and object changes, the compositionality issue of the pre-trained models, which often cannot generate compositions of two or more objects or attributes, is another problem, making our method difficult to accommodate changes to the entire layout.

\section{Discussion and Future Work}

Perturb-and-Revise (PnR) is a training-free editing method that is fast and effective, opening up new possibilities. The main point of the paper is that parameter perturbation is of prime importance in achieving these results. This approach can be compared to SDEdit (Meng et al., 2021), which injects Gaussian noise for image editing and is commonly adopted in many image editing pipelines. Indeed, PnR demonstrates that similar yet general principles can be applied in parameter space for 3D editing.

While PnR currently focuses on static 3D scenes, future research could extend the proposed methods to 4D neural fields representing dynamic scenes. This extension would enable powerful video editing applications, such as modifying the motion of objects or characters in a 3D-consistent way while preserving their appearance and physical plausibility.

\clearpage

\end{document}